\documentclass{article}

\usepackage{PRIMEarxiv}

\usepackage{amsmath, xcolor}
\usepackage{float}

\usepackage[utf8]{inputenc} 
\usepackage[T1]{fontenc}    
\usepackage{hyperref}       
\usepackage{url}            
\usepackage{amssymb}
\usepackage{algorithm, algpseudocode}
\usepackage{booktabs}       
\usepackage{amsfonts}       
\usepackage{nicefrac}       
\usepackage{microtype}      
\usepackage{lipsum}
\usepackage{fancyhdr}       
\usepackage{graphicx}       
\usepackage{subcaption}
\graphicspath{{media/}}     

\title{Differential Machine Learning for Time Series Prediction

}

\author{
  Akash Yadav \\
  Universitat Pompeu Fabra \\
  Barcelona School of Economics \\
  Ram\'on Trias Fargas 25-27, Barcelona 08005, Spain\\
  \texttt{akash.yadav@bse.eu} \\
  \And
  Eulalia Nualart \\
  Universitat Pompeu Fabra \\
  Barcelona School of Economics \\
  Ram\'on Trias Fargas 25-27, Barcelona 08005, Spain\\
  \texttt{eulalia.nualart@upf.edu}
}

\begin{document}
\maketitle

\begin{abstract}

Accurate time series prediction is challenging due to the inherent nonlinearity and sensitivity to initial conditions. We propose a novel approach that enhances neural network predictions through differential learning, which involves training models on both the original time series and its differential series. Specifically, we develop a differential long short-term memory (Diff-LSTM) network that uses a shared LSTM cell to simultaneously process both data streams, effectively capturing intrinsic patterns and temporal dynamics. Evaluated on the Mackey-Glass, Lorenz, and Rössler chaotic time series, as well as a real-world financial dataset from ACI Worldwide Inc., our results demonstrate that the Diff-LSTM network outperforms prevalent models such as recurrent neural networks, convolutional neural networks, and bidirectional and encoder-decoder LSTM networks in both short-term and long-term predictions. This framework offers a promising solution for enhancing time series prediction, even when comprehensive knowledge of the underlying dynamics of the time series is not fully available.

\end{abstract}


\section{Introduction}

Time series prediction is a fundamental task in various fields, including finance, economics, engineering, and environmental science. Accurately forecasting complex time series remains a significant challenge due to their inherent nonlinearity and sensitivity to initial conditions. To address this, we propose a novel approach that leverages differential learning to enhance the predictive capabilities of neural networks for time series data.

Differential learning, introduced by Huge and Savine \cite{huge2020differential}, involves training a model not only on the original time series data but also on its differential series—the sequence of changes between consecutive data points. This approach enables the model to capture both the levels and the dynamics of change within the data, leading to a more comprehensive understanding of the underlying processes.

We apply this framework to long short-term memory (LSTM) networks, resulting in the development of a new architecture called the differential LSTM (Diff-LSTM) network. LSTM networks \cite{HochSchm97} are a type of recurrent neural network (RNN) designed to capture long-term dependencies in sequential data. An LSTM cell, the fundamental building block of an LSTM network, processes input data at each time step and maintains a hidden state over time.

The Diff-LSTM network utilizes a shared LSTM cell to simultaneously process both the original and differential time series. This shared representation allows the network to learn from the intrinsic patterns of the data as well as the nuances of its temporal dynamics. As demonstrated in \cite{huge2020differential}, this augmentation is not specific to any particular neural network architecture; however, for the tasks at hand, we focus on the LSTM variant.

To evaluate the effectiveness of the Diff-LSTM network, we conduct experiments using the Mackey-Glass, Lorenz, and Rössler time series \cite{MG, LorenzDeterministicNonperiodicFlow, Rossler1976397}, which are well-known chaotic systems often used as benchmarks for time series prediction, alongside the ACI finance series as a real-world dataset. These datasets are chosen to allow for a direct comparison of our results with benchmark models. We compare the performance of the Diff-LSTM network against prevalent models, including recurrent neural networks (RNNs), convolutional neural networks (CNNs), bidirectional LSTM (BD-LSTM), and encoder-decoder LSTM (ED-LSTM) architectures.

A baseline comparison of these models for chaotic time series prediction tasks is provided by Chandra et al. \cite{access2021}. In their baseline models, LSTM networks use 10 hidden units. We have maintained this baseline while comparing our model’s results with those of traditional LSTM networks. However, it is important to note that a direct comparison of these models is not straightforward and depends on the characteristics of the architecture and model training.

Our results demonstrate that the Diff-LSTM network significantly outperforms these traditional models in both short- and long-horizon time series prediction tasks. The incorporation of differential learning leads to improved accuracy in forecasting both the original and differential series, constrained only by the accuracy with which we can estimate these "differentials." The Diff-LSTM network effectively captures the complex dynamics of chaotic time series, showcasing its potential as a powerful tool for time series analysis.

Our approach draws inspiration from Salgado et al. \cite{SALGADO2022498} and their fuzzy derivative model for time series prediction, but instead of using fuzzy logic, we harness the powerful potential of LSTM networks to predict both the original and differential series within a single framework. This approach also enables multi-step ahead predictions for both series.

An important question arises: how is this approach different from multivariate learning through deep neural networks? This is where the customized loss function comes into play. The series generated by these differentials does not merely serve as additional input data. By simultaneously predicting these differentials and incorporating their effect on the loss function, we provide an effective way to enhance learning, especially with small datasets and limited-capacity models.

Recent work on physics-informed neural networks \cite{RAISSI2019686, LIU2023116500} has demonstrated the effectiveness of using theoretical knowledge of natural states to guide the learning process of deep neural networks for improved predictive results. Our work builds on this concept by leveraging the specific learning from the geometry of time series, making necessary adjustments to the loss function for our LSTM model. A better estimate of the differentials significantly improves learning by the LSTM. For example, when applied to ACI Worldwide, Inc. Common Stock data from NASDAQ, with differentials calculated after applying a polynomial fit to the original dataset (e.g., using the Savitzky-Golay filter \cite{savitzky-golay}), the Diff-LSTM produces better predictive results compared to frequently used neural network architectures.

This opens up a new class of applications for such networks, where a technically sound estimate of the differentials can lead to improved predictions of the original time series, even when the underlying processes that govern the time series dynamics are unknown. It is crucial to note the caveat associated with this: the success of the approach depends on how "good" our estimate of these differentials is. If we augment our input data with noise and train our model, a custom loss function can act as a double-edged sword in these instances. However, when the change in the system over time is a crucial quantity to estimate, this framework also provides a one-shot solution to predict both the evolution of the system and its changes over time.

The rest of the paper is organized as follows: Section \ref{sec:NNarchitecture} presents the architecture of the differential LSTM network. Section \ref{sec:applications_results} presents the experiments and results. Finally, Section \ref{sec:conclusion} concludes the paper with a discussion of future work.

\section{Differential LSTM network architecture}
\label{sec:NNarchitecture}

The differential neural network architecture used in this work draws upon the concept proposed by Savine et al. \cite{huge2020differential}, where the feedforward and backward propagation equations are combined into a single network representation, called the twin network. This network corresponds to the computation of both the function values and its derivatives. The network is then trained on a custom loss function that incorporates the mean squared error of the derivatives as well.

While the twin network approach estimates the derivatives of a function through backward propagation, we focus on time series governed by specific differential equations, such as delay differential equations or those derived from polynomial fits to the original time series. This approach ensures a smooth representation of the derivatives, allowing the model to more effectively capture the underlying dynamics of the time series.

Additionally, a modified network architecture using LSTM networks provides a useful contrast in performance for non-stationary time series prediction applications. Long-horizon forecasting of financial time series has shown promising results with LSTM architecture \cite{FinancialAN}. The novelty of the architecture presented in this paper lies in the shared learning approach, which incorporates derivatives of the original time series (differential learning) within an LSTM network framework.

\subsection{Loss Function and differential regularization}
\label{sec: loss_fn}

Training with differential labels, which combines errors in both the values and the derivatives, introduces an additional hyperparameter, $\lambda$. The loss function can be written as:
\begin{equation*}
    \mathcal{L} = \textnormal{MSE} + \lambda \overline{\textnormal{MSE}},
\end{equation*}
where MSE and $\overline{\textnormal{MSE}}$ represent the mean squared error of the values and the derivative values, respectively. Differential training provides a safeguard by tracking the shape of the function at all times, leading to a better fit for the predicted values.

Training on both the original data and its differentials serves as a form of regularization, reducing variance without compromising the model's ability to learn the true underlying patterns. This approach leverages additional information from the differentials to enhance generalization, similar to data augmentation, without constraining the model's capacity under certain conditions, as noted in \cite{huge2020differential}. A mathematical deep dive into properties of differential regularization is provided in \cite{RePEc:arx:papers:2405.01233}.

It is important to note that the relative magnitude of these terms can significantly influence the training process. If one term is naturally larger due to higher variance in its target variable, it can dominate the total loss, causing the model to prioritize minimizing that term over the other. Therefore, scaling the data appropriately becomes crucial. Moreover, the accuracy of our estimate of the differentials significantly impacts the model's sensitivity to changes in the magnitude of this hyperparameter.

\subsection{Differential network architecture}

We propose certain modifications to a standard LSTM network, where both the original time series and the series of differentials are provided as input to a shared LSTM cell. This shared cell learns common weights and biases, promoting the extraction of shared features and capturing temporal dependencies beneficial for predicting both the original series and its differential counterpart.

Separate fully connected dense layers then process the concatenated features to extract task-relevant information. A key point to note is that our architecture uses a linear activation function for these dense layers. As a result, the non-linearity of the time series is captured by the shared LSTM cell itself. This is noteworthy, considering that the number of hidden layers remains the same (in our case, 10 for comparative purposes), yet the model can predict both short-term and long-term evolutions of both the original and differential series. Furthermore, as illustrated in Table \ref{tab:diffLSTMarchitecture}, the total number of parameters is 900, which is considerably fewer than those of more complex architectures like BD-LSTM (1170), ED-LSTM (1331), and CNN (1016), with the number of parameters indicated in parentheses. Crucially, the number of hidden layers, input size, and output size are consistent across all models to ensure a fair comparison.

Our proposed architecture works by adding a layer of inputs and hidden states for the differential values of the time series. This approach is similar to how multi-task learning \cite{multitask_learning} is implemented in recurrent neural networks. By learning the shared weights and biases of the network that minimize the loss function—comprising both the error in predicted values and their differentials—we obtain an improved estimate of how the time series evolves over time.

\begin{table}
\centering
\caption{Differential LSTM architecture.}
\label{tab:diffLSTMarchitecture}
\begin{tabular}{|l|l|l|l|}
\hline
\textbf{Layer (type)} & \textbf{Output Shape} & \textbf{Param \#} & \textbf{Connected to} \\
\hline
Shared LSTM & (None, 10) & 480 & Input Layer , Input Layer Differential \\
\hline
Output Layer (Dense) & (None, 10) & 210 & Shared LSTM \\
\hline
Output Layer Differential (Dense) & (None, 10) & 210 & Shared LSTM \\
\hline
\end{tabular}

\end{table}
Next, we describe the key steps involved in processing and transforming the original and differential inputs, and how these are subsequently used to generate predictions (outputs).

\subsubsection{Forward propagation}

We begin by describing the forward propagation setup for the differential LSTM network, using a notation similar to that in \cite{access2021}. Given an observed time series $x_t$ and its derivative $x'_t$,
we denote the inputs to each shared LSTM cell as $X_t$ for the original time series values and $X'_t$. This can be explicitly expressed as:
\begin{align*}
X_t &= \big(x_t, x_{t+T}, \ldots, x_{t+(D-1)T}\big) \quad \text{corresponding to}\ D\ \text{values of the original series} \\
X'_t &= \big(x'_t, x'_{t+T}, \ldots, x'_{t+(D-2)T}\big) \quad \text{corresponding to}\ D-1\ \text{values of the differential series,} 
\end{align*}
where $T$ is the time delay and $D$ is the embedding dimension (window size). We assume that we observe $x_0, \ldots, x_N$ points in the original series.  Then, $t$
ranges in $\{0,1, \ldots, N- (D-1)T\}$. We assume that we only observe the points $x'_0,\ldots, x'_{N-T}$ of the differential series since we should not use the differential series value that contains information about future values of the original series. This still needs to account for the prediction horizon, which in turn determines the size of the output as demonstrated while computing the loss function below.

To maintain a coherent learned latent representation for both inputs, we use different notation for the hidden states of the original and differential values of the shared LSTM, namely $h_t$ and $\bar{h}_t$. We denote the output predicting the multistep ahead next values and corresponding differentials by $Y_t$ and $\bar{Y}_t$ respectively. That is,
\begin{align*}
Y_t &= \big(y_{t+1}, y_{t+2}, \ldots, y_{t+H}\big) \quad \text{corresponding to}\ H \ \text{predictions for the original series} \\
\bar{Y}_t &= \big(\bar{y}_{t+1}, \bar{y}_{t+2}, \ldots, \bar{y}_{t+H}\big) \quad \text{corresponding to}\ H\ \text{predictions for the differential series} 
\end{align*}
where $H$ is the prediction horizon. 
As we will see in the applications, we make predictions for 1 to 10 steps ahead, meaning \( H = 10 \) in this case, as in \cite{access2021}.  

The shared weights for the input, forget, and output gates, as well as the candidate cell states, are denoted by \( W^i, W^f, W^o, W^g \). Similarly, the shared recurrent weights and biases are represented as \( U^i, U^f, U^o, U^g \) and \( b^i, b^f, b^o, b^g \), respectively. For ease of representation and model specification, we denote the weights and biases for the output layer as \( W^y \) and \( b^y \).  

Using this notation, at each time step \( t \), the LSTM cell computes the hidden state \( h_t \) using the following equations:  
\begin{align*}
    i_t &= \sigma(X_t U^i + h_{t-1} W^i + b^i) \quad &\text{(Input gate)} \\
    f_t &= \sigma(X_t U^f + h_{t-1} W^f + b^f) \quad &\text{(Forget gate)} \\
    o_t &= \sigma(X_t U^o + h_{t-1} W^o + b^o) \quad &\text{(Output gate)} \\
    \tilde{C}_t &= \tanh(X_t U^g + h_{t-1} W^g + b^g) \quad &\text{(Candidate cell state)} \\
    C_t &= f_t \ast C_{t-1} + i_t \ast \tilde{C}_t \quad &\text{(Cell state update)} \\
    h_t &= o_t \ast \tanh(C_t) \quad &\text{(Hidden state update)}
\end{align*}
where \(\sigma\) and \(\tanh\) represent the sigmoid and hyperbolic tangent activation functions, respectively, and \(\ast\) denotes element-wise multiplication. For the differential series, similar equations apply to compute the hidden state \(\bar{h}_t\):  
\begin{align}
\label{eq:LSTM_cell_eq}
\begin{split}
    \bar{i}_t &= \sigma\left(X'_t U^i + \bar{h}_{t-1} W^i\right)  \\
    \bar{f}_t &= \sigma\left(X'_t U^f + \bar{h}_{t-1} W^f\right)  \\
    \bar{o}_t &= \sigma\left(X'_t U^o + \bar{h}_{t-1} W^o\right)  \\
    \tilde{C}_t &= \tanh\left(X'_t U^g + \bar{h}_{t-1} W^g\right)  \\
    \bar{C}_t &= \bar{f}_t \ast \bar{C}_{t-1} + \bar{i}_t \ast \tilde{C}_t  \\
    \bar{h}_t &= \bar{o}_t \ast \tanh(\bar{C}_t).
\end{split}
\end{align}
The initial values are set as \(C_0 = 0\), \(\bar{C}_0 = 0\), \(h_0 = 0\), and \(\bar{h}_0=0\). In our application, all gates have the same dimensions, \(d_h\) and \(d_{\bar{h}}\), which represent the sizes of the hidden states for the original and differential series, respectively. As shown in Table \ref{tab:diffLSTMarchitecture}, both \(d_h\) and \(d_{\bar{h}}\) are set to 10. This condition is maintained to ensure a fair comparison across different models.  

Finally, the outputs for the original and differential series are computed as:  
\begin{equation}
    Y_t = W^yh_t + b^y\qquad  \text{and}\qquad  \bar{Y}_t = W^y\bar{h}_t + b^y, \label{eq:2}
\end{equation}
where \(W^y \in \mathbb{R}^{H \times d_h}\) and \(b^y \in \mathbb{R}^{H}\).

\begin{figure}
    \centering
    \includegraphics[width=1.0\textwidth]{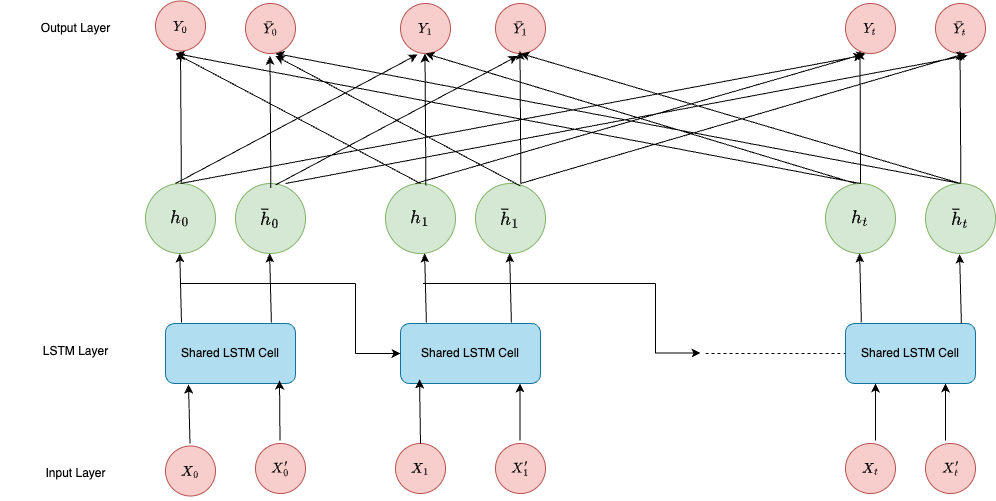} 
    \caption{Differential LSTM architecture where $t = N- (D-1)T$.}
    \label{fig:DiffLSTM}
\end{figure}

The network  architecture is summarized in Figure \ref{fig:DiffLSTM}. 
For the shared LSTM cell, we use the hyperbolic tangent (\(\tanh\)) and sigmoid (\(\sigma\)) activation functions \cite{Gers2000LearningTF}, along with a linear activation function for our dense (fully connected) layers. The choice of these activation functions determines the range of our scaled dataset. To isolate the impact of differential learning on our prediction results, these activation functions are kept consistent across all LSTM cells used in the prevalent architectures.  

Furthermore, the learning of intricate nonlinear relationships primarily occurs through the shared LSTM cell, while the dense layers act as conduits for generating distinct predictions from the original and differential series through simple output transformations. Despite using a more conservative number of parameters compared to BD-LSTM, ED-LSTM, and CNN, our model demonstrates superior predictive performance, as evidenced by the test cases.

\subsubsection{The loss function and backward propagation}
\label{loss_architecture}

Having described the forward propagation of our differential LSTM, detailing how hidden states \( h_t \) and \( \bar{h}_t \) are computed at each time step, we now define the objective that drives the learning process. Specifically, our goal is to find the weights and biases that minimize the loss function, which measures how well the model predicts both the original time series and its differential. By formulating this multistep forecast loss and applying backward propagation over time, we can iteratively update all weights and biases to reduce prediction errors. The remainder of this subsection focuses on explaining this process.

The loss function combines the errors from predictions of both the next values and the next differentials, with a coefficient \( \lambda \) controlling the weight of each loss. This requires careful alignment between the predictions and the original values to compute the squared error terms for both the original and differential series. Since we aim to predict \( H \) steps into the future, the last input window from which we can predict all \( H \) future steps is \( M = N - (D - 1)T - H \). Therefore, the range of \( t \) over which we have the full set of \( H \) predictions extends from \( 0 \) to \( M \), and our loss function is defined as:
$$
\mathcal{L} =  \mathcal{L}_{\text{original}} + \lambda \cdot \mathcal{L}_{\text{differential}},$$
where $\lambda$ is a tunable hyperparameter,
$$
\mathcal{L}_{\text{original}} 
= \frac{1}{M \cdot H} 
\sum_{t=0}^{M} \sum_{k=1}^{H} \bigl(y_{t+k} - x_{t+(D-1)T+k}\bigr)^2,
$$
and
$$
\mathcal{L}_{\text{differential}} 
= \frac{1}{M \cdot H} 
\sum_{t=0}^{M} \sum_{k=1}^{H} \bigl(\bar{y}_{t+k} - x'_{t+(D-2)T+k}\bigr)^2.
$$

Observe that since the length of the original series and its differential counterpart are \( N+1 \) and \( N+1-T \), respectively, it turns out that \( M \) has the same range. It is crucial to note that if \( \lambda = 0 \), then this becomes a classical multivariate learning task.

To minimize the loss function, at each time step \( t \), we first study how the error propagates backward through all \( H \) predicted steps. That is,  
\begin{equation*}\begin{split}
    \frac{\partial \mathcal{L}}{\partial h_t} &= \sum_{k=1}^{H} \frac{\partial \mathcal{L}}{\partial y_{t+k}} (W^y_{k})^{\intercal} \\
    &=\sum_{k=1}^{H} \Big(\frac{2}{M H}(y_{t+k}-x_{t+(D-1)T+k}) (W^y_{k})^{\intercal} + \frac{2\lambda}{M H}(\bar{y}_{t+k}-x'_{t+(D-2)T+k}) (W^y_{k})^{\intercal} \Big),
    \end{split}
\end{equation*}
 where $W^y_k$ denotes the $k$-th row of $W^y\in \mathbb{R}^{H \times d_h}$. Similarly, one can compute $\frac{\partial \mathcal{L}}{\partial \bar{h}_t} = \sum_{k=1}^{H} \frac{\partial \mathcal{L}}{\partial \bar{y}_{t+k}} (W^y_{k})^{\intercal}$.
 
Then, the backward propagation through the given states is given by
\[
\frac{\partial \mathcal{L}}{\partial h_{t-1}} = \frac{\partial \mathcal{L}}{\partial h_t} (U^g)^T \ast (1 - \tanh(C_{t-1})^2)
\quad \text{ and } \quad 
\frac{\partial \mathcal{L}}{\partial \bar{h}_{t-1}} = \frac{\partial \mathcal{L}}{\partial \bar{h}_t} (U^g)^T \ast (1 - \tanh(\bar{C}_{t-1})^2).
\]

Next, using a gradient based optimization method such as Adam (see for instance \cite{Kingma:2014vow}), we update the weights and biases of the LSRM cell as
\begin{align*}
    \theta &\leftarrow \theta - \eta \left( \sum_t \frac{\partial \mathcal{L}}{\partial h_t} \frac{\partial h_t}{\partial \theta} + \sum_t \frac{\partial \mathcal{L}}{\partial \bar{h}_t} \frac{\partial \bar{h}_t}{\partial \theta} \right),
\end{align*}
where $\theta \in \{W^i, U^i, b^i, W^f, U^f, b^f, W^o, U^o, b^o, W^g, U^g, b^g\}$ and
$\eta$ is the learning rate. 
The gradients $\frac{\partial \mathcal{L}}{\partial h_t}$ and $\frac{\partial \mathcal{L}}{\partial \bar{h}_t}$ are computed using the backward propagation through time as explained above, and $\frac{\partial h_t}{\partial \theta}$ and $\frac{\partial \bar{h}_t}{\partial \theta}$ are computed using the LSTM cell equations \ref{eq:LSTM_cell_eq}.
 
For the output layer, weight and bias updates are given by
$$
W^y \leftarrow W^y - \eta\left( \sum_{t=0}^{M} \sum_{k=1}^{H}\frac{\partial \mathcal{L}}{\partial y_{t+k}}\frac{\partial y_{t+k}}{\partial W^y} 
\;+\; \sum_{t=0}^{M} \sum_{k=1}^{H}\frac{\partial \mathcal{L}}{\partial \bar{y}_{t+k}}\frac{\partial \bar{y}_{t+k}}{\partial W^y}\right)
$$
and
$$
b^y \leftarrow b^y - \eta\left( \sum_{t=0}^{M} \sum_{k=1}^{H}\frac{\partial \mathcal{L}}{\partial y_{t+k}} 
\;+\; \sum_{t=0}^{M} \sum_{k=1}^{H}\frac{\partial \mathcal{L}}{\partial \bar{y}_{t+k}}\right).
$$
Finally, these optimized weights and biases can now be substituted into equation \ref{eq:2} to obtain the prediction results for both the original and differential series.

\subsection{Takens' theorem}
\label{sec: Taken}

Takens' embedding theorem \cite{takens1981detecting} plays an important role in chaotic time-series analysis. This theorem provides a method for reconstructing the state space of a dynamical system from a time series of observations. According to Takens' theorem, if the embedding dimension \( D \) is sufficiently large, it is possible to capture the essential features and structure of the underlying dynamical system. More specifically, the theorem states that for a smooth dynamical system with an attractor of dimension \( d \), an embedding dimension \( D > 2d \) is generally sufficient to reconstruct the attractor from the time series data \cite{frazier2004chaos}. To apply this to our test cases, we can use the Grassberger-Procaccia algorithm \cite{PhysRevLett.50.346} to first find the fractal dimension \( d \), and then use the formula above to compute the embedding dimension.

This embedding process involves creating a state-space vector by considering time-delayed versions of the observed time series. The significance of Takens' theorem lies in its ability to provide a way to analyze and understand chaotic systems using time-delay embeddings, which can reveal the underlying dynamics even when the original state variables are not directly accessible.

In the context of chaotic time series analysis, such as with the Mackey-Glass, Lorenz, and Rössler time series, Takens' theorem is instrumental. It allows us to reconstruct the phase space of the chaotic attractor, providing insights into the system's behavior and predictability. Sauer et al. \cite{sauer1991embedology} were able to generalize Takens' theorem extending the applicability to systems with fractal, non-integer dimensional attractors. This generalization reduces the required embedding dimension based on the fractal dimension and relaxes the smoothness and generic conditions on the observation function.

This increases the scope of its applicability to real-world datasets, like the ACI Worldwide stock price data. Computing the embedding dimension for such time series becomes more nuanced; however, there are algorithmic methods, such as computing false nearest neighbors \cite{RevModPhys.65.1331}, which work well. When embedding a time series into a \( D \)-dimensional space, points that appear close ("neighbors") may only seem so due to the projection into lower dimensions. If these neighbors separate significantly when the embedding dimension is increased to \( D+1 \), they are considered false neighbors. By iteratively increasing the embedding dimension and measuring the percentage of false neighbors, the optimal embedding dimension is reached when the percentage drops below a small threshold.

We will use Takens' theorem to determine the input window size of our original series. The embedding dimension represents the minimal number of past observations needed to capture the system's state fully, and by equating it to the input window size, we ensure that the LSTM receives enough information to reconstruct the current state of the system, which is essential for accurate prediction. Since the differentials are nothing but linear combinations of the elements of a delay embedding, the dimension of the reconstructed state space in differential coordinates must be the same as in delay coordinates.

The window size for our differential inputs is predetermined by the constraint on the window size of our original series. Since we estimate the differentials at each time step, including more input points from the differential series would effectively amount to using future data to make predictions, which would no longer be a true prediction.

\section{Applications and Results}
\label{sec:applications_results}

\subsection{Chaotic time series}
\label{sec:MG}

\subsubsection{Mackey-Glass time series}

One of the standard benchmarks for assessing time series prediction algorithms is the Mackey-Glass chaotic time series. It is used to model various physiological systems, including blood production, and exhibits a rich array of dynamics, ranging from periodic to chaotic behavior, depending on the parameter settings, see \cite{MG}.

The Mackey-Glass time series is characterized by its nonlinear and chaotic dynamics. Due to its complexity, it has become an important test case for evaluating the effectiveness of various predictive models, including traditional statistical methods, neural networks, and machine learning algorithms. Unlike the other two chaotic systems discussed below [\ref{sec: Lorenz}, \ref{sec: Rossler}], the Mackey-Glass system is governed by a delay differential equation:
\begin{equation*}
    f(x_t) \equiv x'_t = \beta x_t + \frac{\alpha x_{t-\tau}}{1+ x_{t-\tau}^c},
\end{equation*}
where $c = 10$, $\alpha = 0.2$, $\beta = 0.1$ and $\tau = 17$. This time series is chaotic for this choice of parameters.
The upper order time derivatives of the state variable \(x_t\) can be defined recursively as:
\begin{equation*}
    x''_t = f'_{x_t}x'_t + f'_{x_{t-\tau}}x'_{t-\tau}
\end{equation*}
and
\begin{equation*}
x'''_t = f'_{x_t}x''_t + f''_{x_{t-\tau}}\left(x'_{t-\tau}\right)^2 + f'_{x_{t-\tau}}x''_{t-\tau},
\end{equation*}
where $f'_{x_t}$ and $f'_{x_{t-\tau}}$ denote the partial derivatives of $f$ with respect to $x_t$ and $x_{t-\tau}$, respectively.

The numerical solution of the above differential equation can be obtained using the fourth-order Runge-Kutta method with a time step of \( 0.1 \), an initial condition of \( x_0 = 1.2 \), and the assumption that \( x_t = 0 \) for \( t < 0 \).  

\begin{figure}
    \centering
    \includegraphics[width=0.6\textwidth]{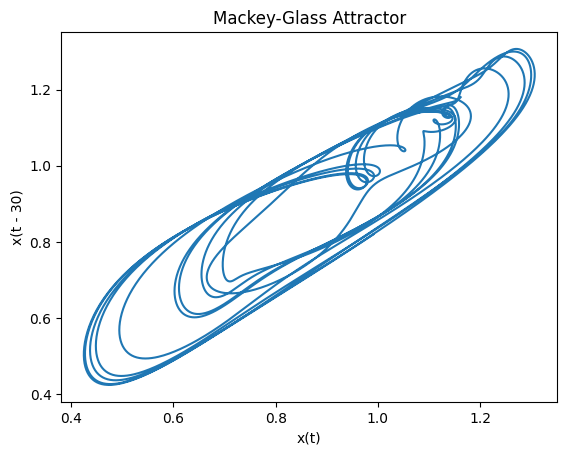} 
    \caption{Chaotic attractors for the Mackey-Glass time series.}
    \label{fig:MGattractor}
\end{figure}

The Mackey-Glass time series also serves as an excellent example for demonstrating the application of Takens' theorem. A reconstruction of the chaotic attractor for the Mackey-Glass time series in two dimensions is shown in Figure \ref{fig:MGattractor}.  

For our analysis, the Mackey-Glass time series is processed into a state-space vector using an embedding dimension of \( D = 5 \), a time lag of \( T = 1 \), and a prediction horizon of \( H = 10 \). The fractal dimension \cite{Grebogifractaldimension} of the Mackey-Glass time series is approximately \( 2.1 \). Applying Takens' theorem, we derive the embedding dimension. The choice of these parameters ensures that the reconstructed phase space accurately captures the dynamics of the original system, facilitating effective time series prediction and analysis.  

Takens' theorem provides the theoretical foundation for this approach, ensuring that the reconstructed dynamics faithfully represent the original chaotic system. By leveraging these principles, we can explore the predictability and complexity of the Mackey-Glass time series within the reconstructed state space.  

To ensure comparability with the baseline models used in \cite{access2021}, we use the same Mackey-Glass time series dataset, augmented by calculating Mackey-Glass differentials based on the underlying delay differential equation. The dataset is then scaled to the range \([-0.5, 0.5]\) using min-max scaling, which serves two purposes. First, this range aligns well with the input range of the hyperbolic tangent activation function, preventing exploding gradients. Second, inverse scaling facilitates direct comparison with the quoted RMSE values in \cite{access2021}.  

For differential regularization, we set \(\lambda = 1\), which strikes a balance between predictive performance for both the original and differential series.

\subsubsection{Lorenz system}
\label{sec: Lorenz}

Next, we consider the Lorenz system \cite{LorenzDeterministicNonperiodicFlow}, a classical example of chaotic dynamics. It consists of a set of three coupled ordinary differential equations \((x_t, g_t, z_t)\). Due to its sensitive dependence on initial conditions, the Lorenz system is a cornerstone of chaos theory. The equations governing the system are:  
\begin{align*}  
    x'_t &= \sigma (g_t - x_t) \\  
    g'_t &= x_t(\rho - z_t) - g_t \\  
    z'_t &= x_t g_t - \beta z_t ,
\end{align*}  
where \(\sigma\), \(\rho\), and \(\beta\) are dimensionless parameters, typically set to \(\sigma = 10\), \(\rho = 28\), and \(\beta = 8/3\).  

For prediction, we select the \( x \)-coordinate of the Lorenz time series and generate 1000 samples using fourth-order Runge-Kutta discretization. To estimate the differentials, the corresponding \( g \) values are also computed. The fractal dimension of the Lorenz system is approximately \( 2.06 \) \cite{Grebogifractaldimension}. Thus, we use an embedding dimension of \( D = 5 \) to fully unfold the attractor in a delay-embedding reconstruction, with a time lag of \( T = 1 \) and a prediction horizon of \( H = 10 \).  
For consistency with the Mackey-Glass case, the data is scaled to the range \([-0.5, 0.5]\). The regularization parameter \(\lambda\) is set to 1, ensuring effective prediction of both the original and differential series.

\subsubsection{Rössler system}
\label{sec: Rossler}

The Rössler system, introduced by Otto Rössler in 1976 \cite{Rossler1976397}, provides another three-dimensional chaotic system and serves as our third benchmark test. While it is often considered simpler in form compared to the Lorenz system, it still exhibits rich chaotic dynamics. The Rössler equations are given by:  

\begin{align*}  
    x'_t &= -g_t - z_t \\  
    g'_t &= x_t + a g_t \\  
    z'_t &= b + (x_t - c)z_t,  
\end{align*}  

where \( a = 0.15 \), \( b = 0.20 \), and \( c = 10 \). Based on these parameter values, the fractal dimension is approximately \( 2.01 \) \cite{Kaplan1979ChaoticBO}. Consequently, we use the same embedding dimension and data preparation approach as in our previous cases. The \( x \)-coordinate is once again selected for prediction, with a 60-40 train-test split, as in the previous experiments. The dataset is generated using the fourth-order Runge-Kutta method.  

\begin{figure}
    \centering
    \includegraphics[width=0.8\textwidth]{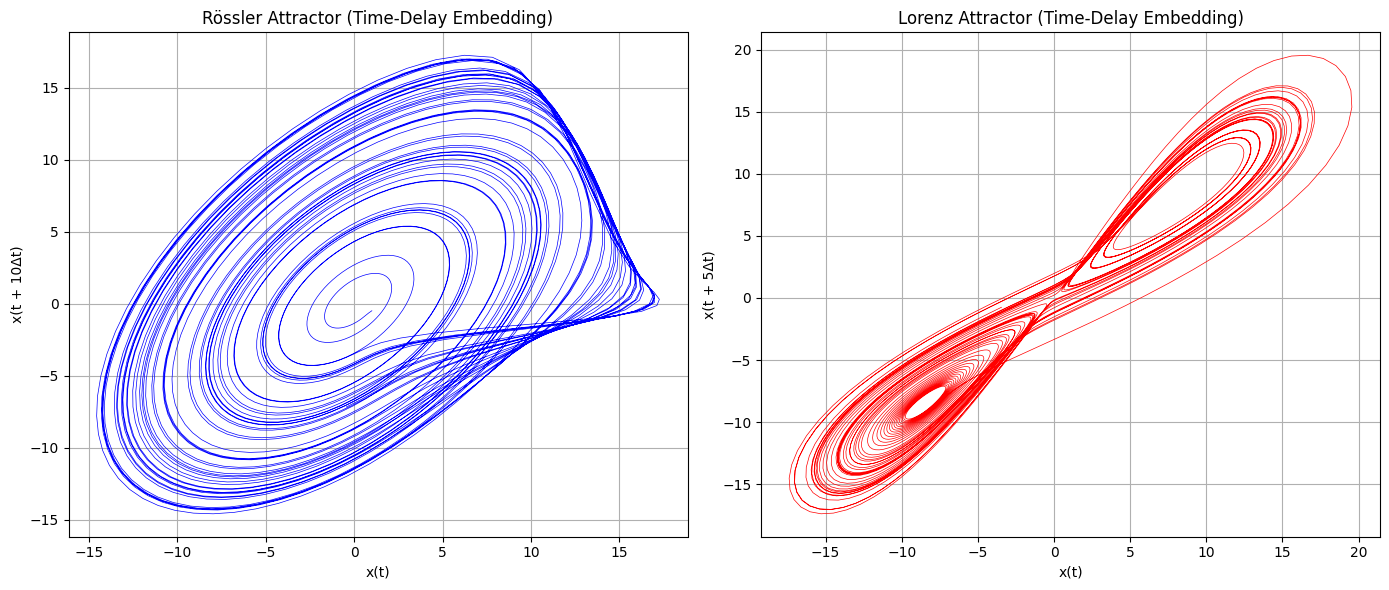} 
    \caption{Chaotic attractors for the Rössler and Lorenz systems.}
    \label{fig:Ross_Lor_attractor}
\end{figure}

Since we focus only on the \( x \)-coordinate for both the Lorenz and Rössler systems, similar to the Mackey-Glass case, we can visualize the reconstruction of chaotic attractors for the Lorenz and Rössler time series in two dimensions using time-delay embedding, as shown in Figure \ref{fig:Ross_Lor_attractor}. Data scaling and the regularization parameter \( \lambda \) follow the same structure as in our previous test cases.  

\subsection{Results and discussion for chaotic time series}
\label{sec:simulated_results}

In this section, we compare the performance of our Diff-LSTM networks with recurrent neural networks (RNNs), convolutional neural networks (CNNs), bidirectional LSTM (BD-LSTM), encoder-decoder LSTM (ED-LSTM), and classical LSTM, using the Mackey-Glass, Lorenz, and Rössler chaotic time series introduced in Section 3.1.  

\begin{figure} 
    \centering
    \caption{RMSE across 10 prediction horizons}
    \begin{subfigure}[c]{0.49\textwidth}
        \centering
        \includegraphics[width=\linewidth]{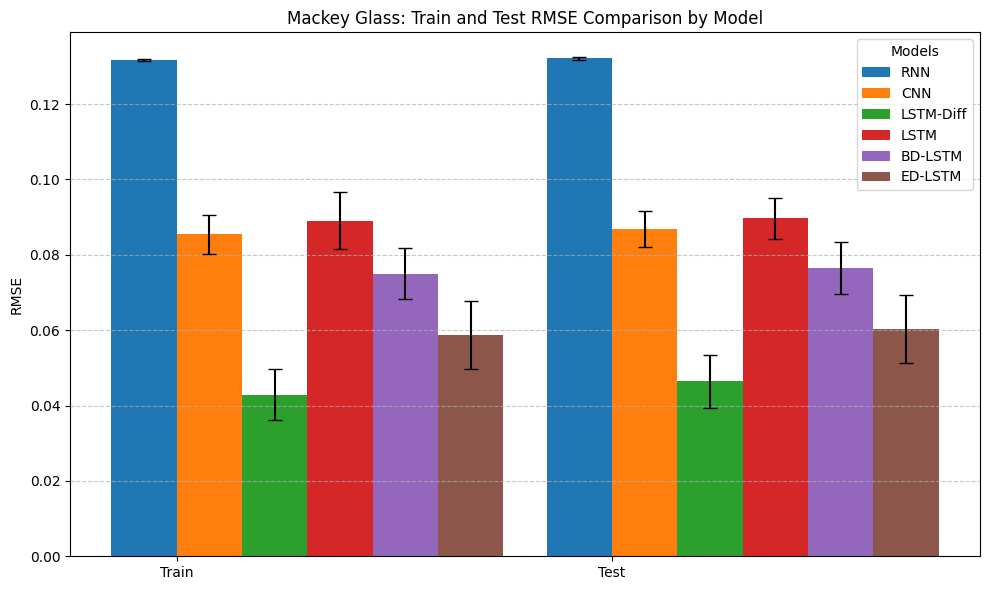}
        \caption{Mackey-Glass series}
        \label{fig:Train_Test_RMSE}
    \end{subfigure}
    \hfill
    \begin{subfigure}[c]{0.49\textwidth}
        \centering
        \includegraphics[width=\linewidth]{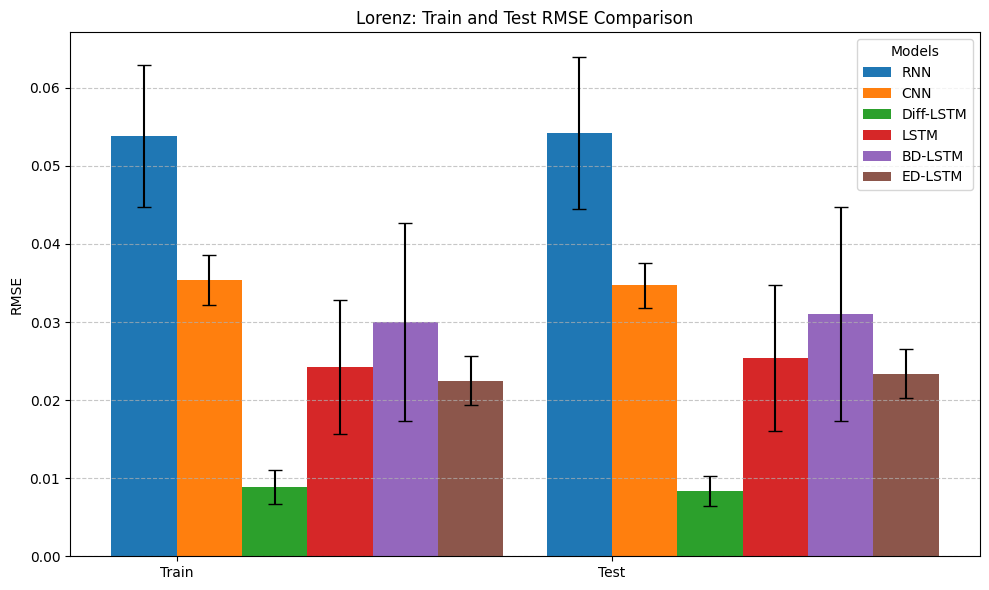}
        \caption{Lorenz series}
        \label{fig:Lorenz_Train_Test_RMSE}
    \end{subfigure}
    \hfill
    \begin{subfigure}[c]{0.49\textwidth}
        \centering
        \includegraphics[width=\linewidth]{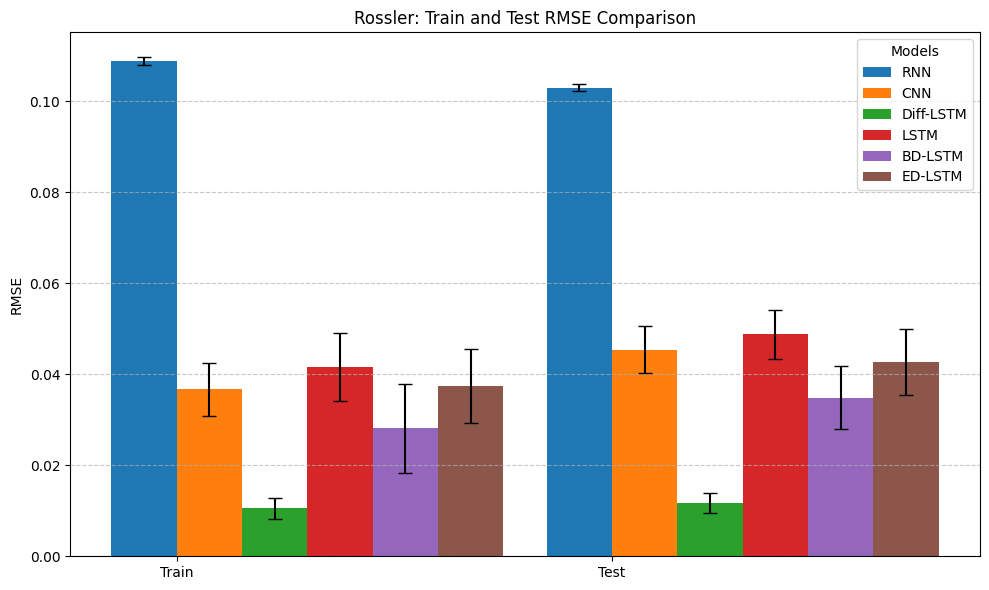}
        \caption{Rössler series}
        \label{fig:Rossler_Train_Test_RMSE}
    \end{subfigure}
    \label{fig4}
\end{figure}

\begin{figure}
    \centering
    \caption{1 to 10 step ahead prediction}
    \label{fig: 1-10 step ahead prediction all benchmark}
    \begin{subfigure}[c]{0.7\textwidth}
        \centering
        \includegraphics[width=\linewidth]{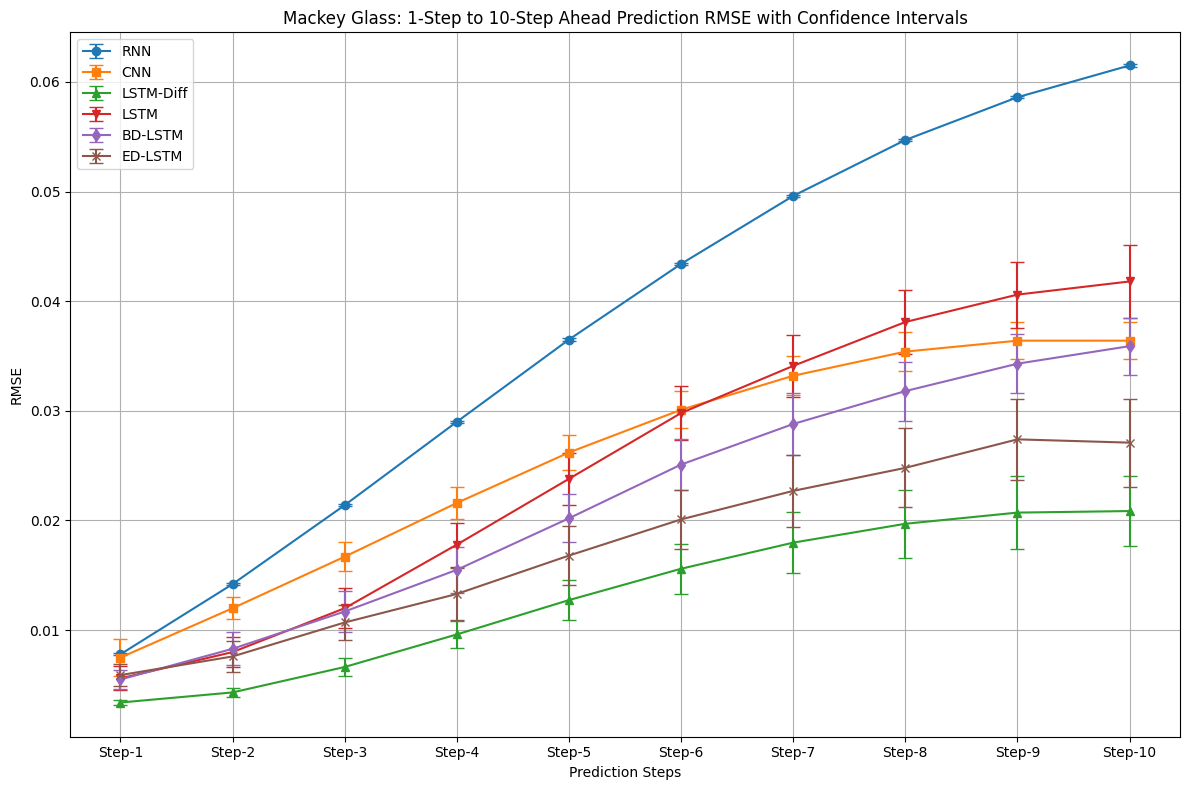}
        \caption{Mackey-Glass series}
        \label{fig:10_step_prediction}
    \end{subfigure}
    \hfill
    \begin{subfigure}[c]{0.7\textwidth}
        \centering
        \includegraphics[width=\linewidth]{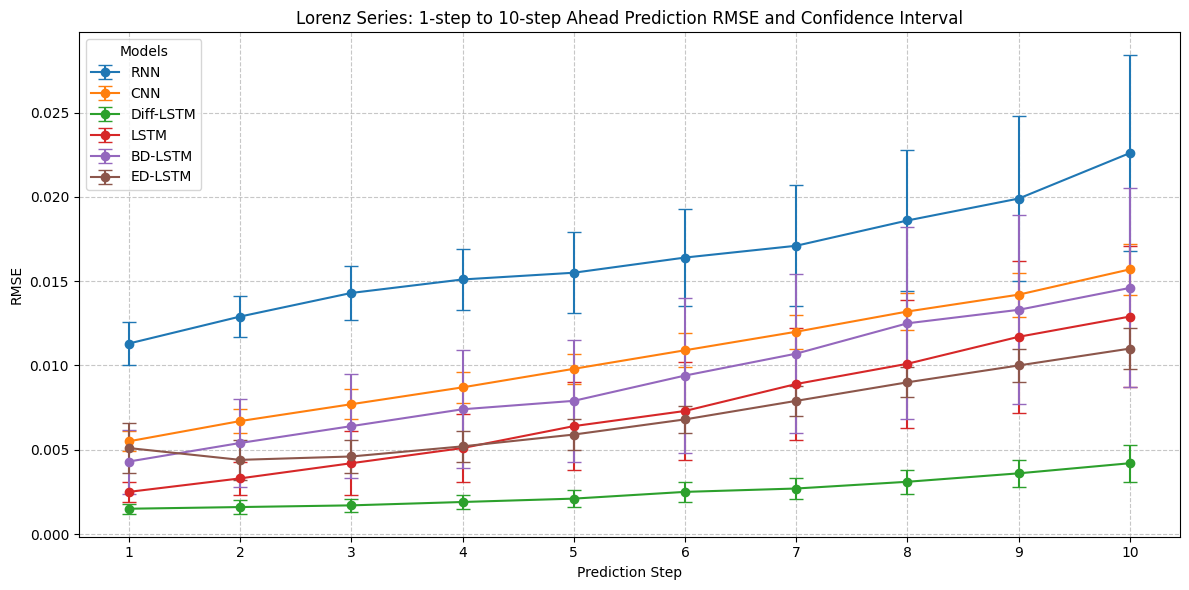}
        \caption{Lorenz series}
        \label{fig:Lorenz_10_step_prediction}
    \end{subfigure}
    \hfill
    \begin{subfigure}[c]{0.7\textwidth}
        \centering
        \includegraphics[width=\linewidth]{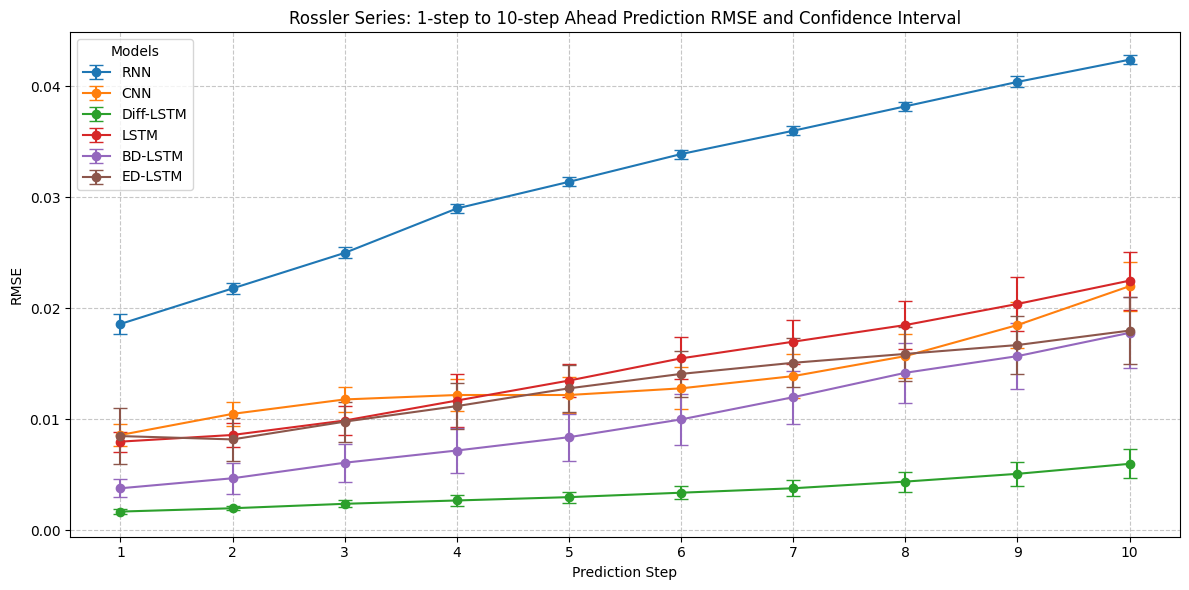}
        \caption{Rössler series}
        \label{fig:Rossler_10_step_prediction}
    \end{subfigure}
    \label{fig5}
\end{figure}

Tables \ref{tab:MG_Diff_LSTM_OG_pred}, \ref{tab:Lorenz_Diff_LSTM_OG_pred}, and \ref{tab:Rossler_Diff_LSTM_OG_pred} present the mean and 95\% confidence interval of the root mean squared error (RMSE) for each prediction horizon across six different architectures. These results are computed for both the training and test datasets of the three benchmark series, based on 30 experimental runs (illustrated in Figure \ref{fig4}). Additionally, one to ten step ahead forecasts are depicted in Figure \ref{fig5}.  

\begin{table}[h]
\centering
\caption{Mean and 95\% confidence interval of the RMSE for the  Mackey-Glass series prediction.}
\label{tab:MG_Diff_LSTM_OG_pred}
\resizebox{\textwidth}{!}{
\begin{tabular}{@{}lcccccc@{}}
\toprule
        & \textbf{RNN}            & \textbf{CNN}            & \textbf{Diff-LSTM}      & \textbf{LSTM}           & \textbf{BD-LSTM}        & \textbf{ED-LSTM}        \\ \midrule
\textbf{Train} & 0.1317 ± 0.0003 & 0.0854 ± 0.0052 & \textbf{0.0428 ± 0.0068}          & 0.0890 ± 0.0076 & 0.0750 ± 0.0067 & 0.0587 ± 0.0090 \\
\textbf{Test}  & 0.1321 ± 0.0003 & 0.0868 ± 0.0048 & \textbf{0.0464 ± 0.0070}          & 0.0897 ± 0.0054 & 0.0765 ± 0.0068 & 0.0602 ± 0.0090 \\
\textbf{Step-1} & 0.0078 ± 0.0001 & 0.0075 ± 0.0007 & \textbf{0.0034 ± 0.0002}          & 0.0056 ± 0.0011 & 0.0055 ± 0.0009 & 0.0059 ± 0.0010 \\
\textbf{Step-2} & 0.0142 ± 0.0001 & 0.0120 ± 0.0010 & \textbf{0.0043 ± 0.0004}          & 0.0080 ± 0.0014 & 0.0083 ± 0.0015 & 0.0076 ± 0.0014 \\
\textbf{Step-3} & 0.0214 ± 0.0001 & 0.0167 ± 0.0013 & \textbf{0.0066 ± 0.0008}          & 0.0120 ± 0.0018 & 0.0117 ± 0.0019 & 0.0107 ± 0.0016 \\
\textbf{Step-4} & 0.0290 ± 0.0001 & 0.0216 ± 0.0015 & \textbf{0.0096 ± 0.0012}          & 0.0178 ± 0.0020 & 0.0155 ± 0.0021 & 0.0133 ± 0.0024 \\
\textbf{Step-5} & 0.0365 ± 0.0001 & 0.0262 ± 0.0016 & \textbf{0.0127 ± 0.0018}          & 0.0238 ± 0.0024 & 0.0202 ± 0.0022 & 0.0168 ± 0.0027 \\
\textbf{Step-6} & 0.0434 ± 0.0001 & 0.0301 ± 0.0017 & \textbf{0.0156 ± 0.0023}          & 0.0298 ± 0.0025 & 0.0251 ± 0.0023 & 0.0201 ± 0.0027 \\
\textbf{Step-7} & 0.0496 ± 0.0001 & 0.0332 ± 0.0018 & \textbf{0.0180 ± 0.0028}          & 0.0341 ± 0.0028 & 0.0288 ± 0.0028 & 0.0227 ± 0.0033 \\
\textbf{Step-8} & 0.0547 ± 0.0001 & 0.0354 ± 0.0018 & \textbf{0.0197 ± 0.0031}          & 0.0381 ± 0.0029 & 0.0318 ± 0.0027 & 0.0248 ± 0.0036 \\
\textbf{Step-9} & 0.0586 ± 0.0001 & 0.0364 ± 0.0017 & \textbf{0.0207 ± 0.0033}          & 0.0406 ± 0.0030 & 0.0343 ± 0.0027 & 0.0274 ± 0.0037 \\
\textbf{Step-10} & 0.0615 ± 0.0001 & 0.0364 ± 0.0017 & \textbf{0.0209 ± 0.0032}         & 0.0418 ± 0.0033 & 0.0359 ± 0.0026 & 0.0271 ± 0.0040 \\ \bottomrule
\end{tabular}
}
\end{table}

\begin{table}[h]
\centering
\caption{Mean and 95\% confidence interval of the RMSE for the  Lorenz series prediction.}
\label{tab:Lorenz_Diff_LSTM_OG_pred}
\resizebox{\textwidth}{!}{
\begin{tabular}{@{}lcccccc@{}}
\toprule
        & \textbf{RNN}            & \textbf{CNN}            & \textbf{Diff-LSTM}                     & \textbf{LSTM}           & \textbf{BD-LSTM}        & \textbf{ED-LSTM}        \\ \midrule
\textbf{Train}   & 0.0538 $\pm$ 0.0091     & 0.0354 $\pm$ 0.0032     & \textbf{0.0089 $\pm$ 0.0022}           & 0.0242 $\pm$ 0.0086     & 0.0300 $\pm$ 0.0127     & 0.0225 $\pm$ 0.0031     \\
\textbf{Test}    & 0.0542 $\pm$ 0.0097     & 0.0347 $\pm$ 0.0029     & \textbf{0.0084 $\pm$ 0.0019}           & 0.0254 $\pm$ 0.0093     & 0.0310 $\pm$ 0.0137     & 0.0234 $\pm$ 0.0031     \\
\textbf{Step-1}  & 0.0113 $\pm$ 0.0013     & 0.0055 $\pm$ 0.0006     & \textbf{0.0015 $\pm$ 0.0003}           & 0.0025 $\pm$ 0.0006     & 0.0043 $\pm$ 0.0019     & 0.0051 $\pm$ 0.0015     \\
\textbf{Step-2}  & 0.0129 $\pm$ 0.0012     & 0.0067 $\pm$ 0.0007     & \textbf{0.0016 $\pm$ 0.0004}           & 0.0033 $\pm$ 0.0010     & 0.0054 $\pm$ 0.0026     & 0.0044 $\pm$ 0.0012     \\
\textbf{Step-3}  & 0.0143 $\pm$ 0.0016     & 0.0077 $\pm$ 0.0009     & \textbf{0.0017 $\pm$ 0.0004}           & 0.0042 $\pm$ 0.0019     & 0.0064 $\pm$ 0.0031     & 0.0046 $\pm$ 0.0010     \\
\textbf{Step-4}  & 0.0151 $\pm$ 0.0018     & 0.0087 $\pm$ 0.0009     & \textbf{0.0019 $\pm$ 0.0004}           & 0.0051 $\pm$ 0.0020     & 0.0074 $\pm$ 0.0035     & 0.0052 $\pm$ 0.0009     \\
\textbf{Step-5}  & 0.0155 $\pm$ 0.0024     & 0.0098 $\pm$ 0.0009     & \textbf{0.0021 $\pm$ 0.0005}           & 0.0064 $\pm$ 0.0026     & 0.0079 $\pm$ 0.0036     & 0.0059 $\pm$ 0.0009     \\
\textbf{Step-6}  & 0.0164 $\pm$ 0.0029     & 0.0109 $\pm$ 0.0010     & \textbf{0.0025 $\pm$ 0.0006}           & 0.0073 $\pm$ 0.0029     & 0.0094 $\pm$ 0.0046     & 0.0068 $\pm$ 0.0008     \\
\textbf{Step-7}  & 0.0171 $\pm$ 0.0036     & 0.0120 $\pm$ 0.0010     & \textbf{0.0027 $\pm$ 0.0006}           & 0.0089 $\pm$ 0.0033     & 0.0107 $\pm$ 0.0047     & 0.0079 $\pm$ 0.0009     \\
\textbf{Step-8}  & 0.0186 $\pm$ 0.0042     & 0.0132 $\pm$ 0.0011     & \textbf{0.0031 $\pm$ 0.0007}           & 0.0101 $\pm$ 0.0038     & 0.0125 $\pm$ 0.0057     & 0.0090 $\pm$ 0.0009     \\
\textbf{Step-9}  & 0.0199 $\pm$ 0.0049     & 0.0142 $\pm$ 0.0013     & \textbf{0.0036 $\pm$ 0.0008}           & 0.0117 $\pm$ 0.0045     & 0.0133 $\pm$ 0.0056     & 0.0100 $\pm$ 0.0010     \\
\textbf{Step-10} & 0.0226 $\pm$ 0.0058     & 0.0157 $\pm$ 0.0015     & \textbf{0.0042 $\pm$ 0.0011}           & 0.0129 $\pm$ 0.0042     & 0.0146 $\pm$ 0.0059     & 0.0110 $\pm$ 0.0012     \\
\bottomrule
\end{tabular}
}
\end{table}

\begin{table}[h]
\centering
\caption{Mean and 95\% confidence interval of the RMSE for the  Rössler series prediction.}
\label{tab:Rossler_Diff_LSTM_OG_pred}
\resizebox{\textwidth}{!}{
\begin{tabular}{@{}lcccccc@{}}
\toprule
        & \textbf{RNN}            & \textbf{CNN}            & \textbf{Diff-LSTM}      & \textbf{LSTM}           & \textbf{BD-LSTM}        & \textbf{ED-LSTM}        \\ \midrule
\textbf{Train}   & 0.1088 $\pm$ 0.0009     & 0.0367 $\pm$ 0.0059     & \textbf{0.0106 $\pm$ 0.0023} & 0.0416 $\pm$ 0.0074     & 0.0281 $\pm$ 0.0098     & 0.0374 $\pm$ 0.0082     \\
\textbf{Test}    & 0.1030 $\pm$ 0.0008     & 0.0454 $\pm$ 0.0052     & \textbf{0.0118 $\pm$ 0.0022} & 0.0488 $\pm$ 0.0054     & 0.0349 $\pm$ 0.0070     & 0.0427 $\pm$ 0.0072     \\
\textbf{Step-1}  & 0.0186 $\pm$ 0.0009     & 0.0086 $\pm$ 0.0010     & \textbf{0.0017 $\pm$ 0.0002} & 0.0080 $\pm$ 0.0009     & 0.0038 $\pm$ 0.0008     & 0.0085 $\pm$ 0.0025     \\
\textbf{Step-2}  & 0.0218 $\pm$ 0.0005     & 0.0105 $\pm$ 0.0011     & \textbf{0.0020 $\pm$ 0.0002} & 0.0086 $\pm$ 0.0011     & 0.0047 $\pm$ 0.0014     & 0.0082 $\pm$ 0.0019     \\
\textbf{Step-3}  & 0.0250 $\pm$ 0.0005     & 0.0118 $\pm$ 0.0011     & \textbf{0.0024 $\pm$ 0.0003} & 0.0099 $\pm$ 0.0013     & 0.0061 $\pm$ 0.0017     & 0.0098 $\pm$ 0.0018     \\
\textbf{Step-4}  & 0.0290 $\pm$ 0.0004     & 0.0122 $\pm$ 0.0014     & \textbf{0.0027 $\pm$ 0.0005} & 0.0117 $\pm$ 0.0014     & 0.0072 $\pm$ 0.0020     & 0.0112 $\pm$ 0.0021     \\
\textbf{Step-5}  & 0.0314 $\pm$ 0.0004     & 0.0122 $\pm$ 0.0016     & \textbf{0.0030 $\pm$ 0.0005} & 0.0135 $\pm$ 0.0015     & 0.0084 $\pm$ 0.0021     & 0.0128 $\pm$ 0.0021     \\
\textbf{Step-6}  & 0.0339 $\pm$ 0.0004     & 0.0128 $\pm$ 0.0019     & \textbf{0.0034 $\pm$ 0.0006} & 0.0155 $\pm$ 0.0019     & 0.0100 $\pm$ 0.0023     & 0.0141 $\pm$ 0.0021     \\
\textbf{Step-7}  & 0.0360 $\pm$ 0.0004     & 0.0139 $\pm$ 0.0020     & \textbf{0.0038 $\pm$ 0.0007} & 0.0170 $\pm$ 0.0020     & 0.0120 $\pm$ 0.0024     & 0.0151 $\pm$ 0.0022     \\
\textbf{Step-8}  & 0.0382 $\pm$ 0.0004     & 0.0157 $\pm$ 0.0020     & \textbf{0.0044 $\pm$ 0.0009} & 0.0185 $\pm$ 0.0022     & 0.0142 $\pm$ 0.0027     & 0.0159 $\pm$ 0.0024     \\
\textbf{Step-9}  & 0.0404 $\pm$ 0.0005     & 0.0185 $\pm$ 0.0021     & \textbf{0.0051 $\pm$ 0.0011} & 0.0204 $\pm$ 0.0024     & 0.0157 $\pm$ 0.0030     & 0.0167 $\pm$ 0.0026     \\
\textbf{Step-10} & 0.0424 $\pm$ 0.0004     & 0.0220 $\pm$ 0.0022     & \textbf{0.0060 $\pm$ 0.0013} & 0.0225 $\pm$ 0.0026     & 0.0178 $\pm$ 0.0032     & 0.0180 $\pm$ 0.0030     \\
\bottomrule
\end{tabular}
}
\end{table}

Our observations indicate that Diff-LSTM significantly outperforms the other deep neural network architectures used for time series forecasting. When applied to the same dataset, with an identical number of hidden layers, Diff-LSTM achieves superior predictive performance while utilizing fewer parameters than other complex networks.  

For one-step-ahead prediction in the Mackey-Glass series, our results are closely aligned with those reported by Salgado et al. \cite{SALGADO2022498}, who employed their fuzzy derivative model ODE-DFS (distributed fuzzy system). The rule-based membership function methodology in \cite{SALGADO2022498} for approximating derivatives stands in stark contrast to the "black-box" nature of an LSTM cell. However, in both approaches, the inclusion of derivative inputs enhances the simultaneous prediction of both the original and differential series with high accuracy. Notably, our NN-based architecture, leveraging LSTM cells, facilitates long-horizon forecasting more effectively.  

While CNN outperforms LSTM in predicting both the Mackey-Glass and Rössler series, Diff-LSTM achieves the best overall performance across all three benchmark tests.

Next, in Tables \ref{tab: MG_Differential_series_results}, \ref{tab: Lorenz_Differential_series_results}, and \ref{tab: Rossler_Differential_series_results}, we present a summary of the statistical performance of Diff-LSTM in predicting the three differential time series studied above. Ensuring that both the original and differential series predictions remain consistent encourages the network to develop an internal representation that more accurately captures the underlying system dynamics. This is because the differential channel quickly reveals any discrepancies between the predicted trajectory and its local slope.

  \begin{table}
\centering
\caption{Mackey-Glass differential series prediction}
\label{tab: MG_Differential_series_results}
\begin{tabular}{lcccccc}
\toprule
\textbf{} & \textbf{Mean} & \textbf{SD} & \textbf{CI\_LB} & \textbf{CI\_UB} & \textbf{Min} & \textbf{Max} \\
\midrule
\textbf{Train RMSE} & 0.00816 & 0.00282 & 0.00716 & 0.00916 & 0.00586 & 0.01807 \\
\textbf{Test RMSE} & 0.00874 & 0.00282 & 0.00774 & 0.00974 & 0.00661 & 0.01888 \\
\textbf{Step-1}     & 0.00109 & 0.00010 & 0.00106 & 0.00112 & 0.00101 & 0.00143 \\
\textbf{Step-2}     & 0.00146 & 0.00039 & 0.00129 & 0.00163 & 0.00114 & 0.00272 \\
\textbf{Step-3}     & 0.00217 & 0.00065 & 0.00194 & 0.00240 & 0.00165 & 0.00425 \\
\textbf{Step-4}     & 0.00290 & 0.00088 & 0.00255 & 0.00325 & 0.00224 & 0.00597 \\
\textbf{Step-5}     & 0.00334 & 0.00102 & 0.00297 & 0.00371 & 0.00257 & 0.00706 \\
\textbf{Step-6}     & 0.00349 & 0.00112 & 0.00309 & 0.00389 & 0.00264 & 0.00775 \\
\textbf{Step-7}     & 0.00342 & 0.00111 & 0.00302 & 0.00382 & 0.00256 & 0.00758 \\
\textbf{Step-8}     & 0.00319 & 0.00118 & 0.00276 & 0.00362 & 0.00237 & 0.00704 \\
\textbf{Step-9}     & 0.00284 & 0.00106 & 0.00247 & 0.00321 & 0.00208 & 0.00643 \\
\textbf{Step-10}    & 0.00254 & 0.00095 & 0.00219 & 0.00289 & 0.00186 & 0.00590 \\
\bottomrule
\end{tabular}
\end{table}

\begin{table}
\centering
\caption{Lorenz differential series prediction}
\label{tab: Lorenz_Differential_series_results}
\begin{tabular}{lcccccc}
\toprule
\textbf{} & \textbf{Mean} & \textbf{SD} & \textbf{CI\_LB} & \textbf{CI\_UB} & \textbf{Min} & \textbf{Max} \\
\midrule
\textbf{Train RMSE} & 0.02091 & 0.01675 & 0.01492 & 0.02690 & 0.00932 & 0.09793 \\
\textbf{Test RMSE}  & 0.02008 & 0.01338 & 0.01529 & 0.02486 & 0.01242 & 0.08472 \\
\textbf{Step-1}     & 0.00201 & 0.00084 & 0.00171 & 0.00231 & 0.00092 & 0.00477 \\
\textbf{Step-2}     & 0.00250 & 0.00139 & 0.00201 & 0.00300 & 0.00111 & 0.00803 \\
\textbf{Step-3}     & 0.00340 & 0.00268 & 0.00245 & 0.00436 & 0.00147 & 0.01624 \\
\textbf{Step-4}     & 0.00401 & 0.00239 & 0.00315 & 0.00486 & 0.00204 & 0.01454 \\
\textbf{Step-5}     & 0.00495 & 0.00344 & 0.00372 & 0.00618 & 0.00272 & 0.02172 \\
\textbf{Step-6}     & 0.00583 & 0.00386 & 0.00445 & 0.00721 & 0.00345 & 0.02456 \\
\textbf{Step-7}     & 0.00684 & 0.00486 & 0.00510 & 0.00859 & 0.00422 & 0.03127 \\
\textbf{Step-8}     & 0.00804 & 0.00545 & 0.00609 & 0.00999 & 0.00504 & 0.03494 \\
\textbf{Step-9}     & 0.00931 & 0.00646 & 0.00700 & 0.01163 & 0.00591 & 0.04091 \\
\textbf{Step-10}    & 0.01033 & 0.00659 & 0.00797 & 0.01270 & 0.00670 & 0.04885 \\
\bottomrule
\end{tabular}
\end{table}

\begin{table}
\centering
\caption{Rössler differential series prediction}
\label{tab: Rossler_Differential_series_results}
\begin{tabular}{lcccccc}
\toprule
\textbf{} & \textbf{Mean} & \textbf{SD} & \textbf{CI\_LB} & \textbf{CI\_UB} & \textbf{Min} & \textbf{Max} \\
\midrule
\textbf{Train RMSE} & 0.01830 & 0.01305 & 0.01363 & 0.02297 & 0.00812 & 0.06640 \\
\textbf{Test RMSE}  & 0.01996 & 0.01168 & 0.01578 & 0.02414 & 0.01010 & 0.06177 \\
\textbf{Step-1}     & 0.00352 & 0.00134 & 0.00304 & 0.00400 & 0.00232 & 0.00714 \\
\textbf{Step-2}     & 0.00415 & 0.00173 & 0.00353 & 0.00477 & 0.00249 & 0.01009 \\
\textbf{Step-3}     & 0.00462 & 0.00251 & 0.00372 & 0.00552 & 0.00262 & 0.01396 \\
\textbf{Step-4}     & 0.00544 & 0.00289 & 0.00441 & 0.00648 & 0.00282 & 0.01628 \\
\textbf{Step-5}     & 0.00567 & 0.00254 & 0.00477 & 0.00658 & 0.00317 & 0.01611 \\
\textbf{Step-6}     & 0.00633 & 0.00259 & 0.00540 & 0.00726 & 0.00333 & 0.01509 \\
\textbf{Step-7}     & 0.00620 & 0.00309 & 0.00510 & 0.00731 & 0.00327 & 0.01640 \\
\textbf{Step-8}     & 0.00608 & 0.00406 & 0.00463 & 0.00753 & 0.00300 & 0.02197 \\
\textbf{Step-9}     & 0.00740 & 0.00584 & 0.00530 & 0.00949 & 0.00282 & 0.02846 \\
\textbf{Step-10}    & 0.01044 & 0.00708 & 0.00790 & 0.01297 & 0.00449 & 0.03412 \\
\bottomrule
\end{tabular}
\end{table}
We observe that our one-step-ahead prediction results for the Mackey-Glass differential series closely align with those reported in \cite{SALGADO2022498}, demonstrating the effectiveness of our architecture in learning both the original and differential series simultaneously—without incurring additional computational cost.

\begin{figure}
    \centering
    \includegraphics[width=0.6\textwidth]{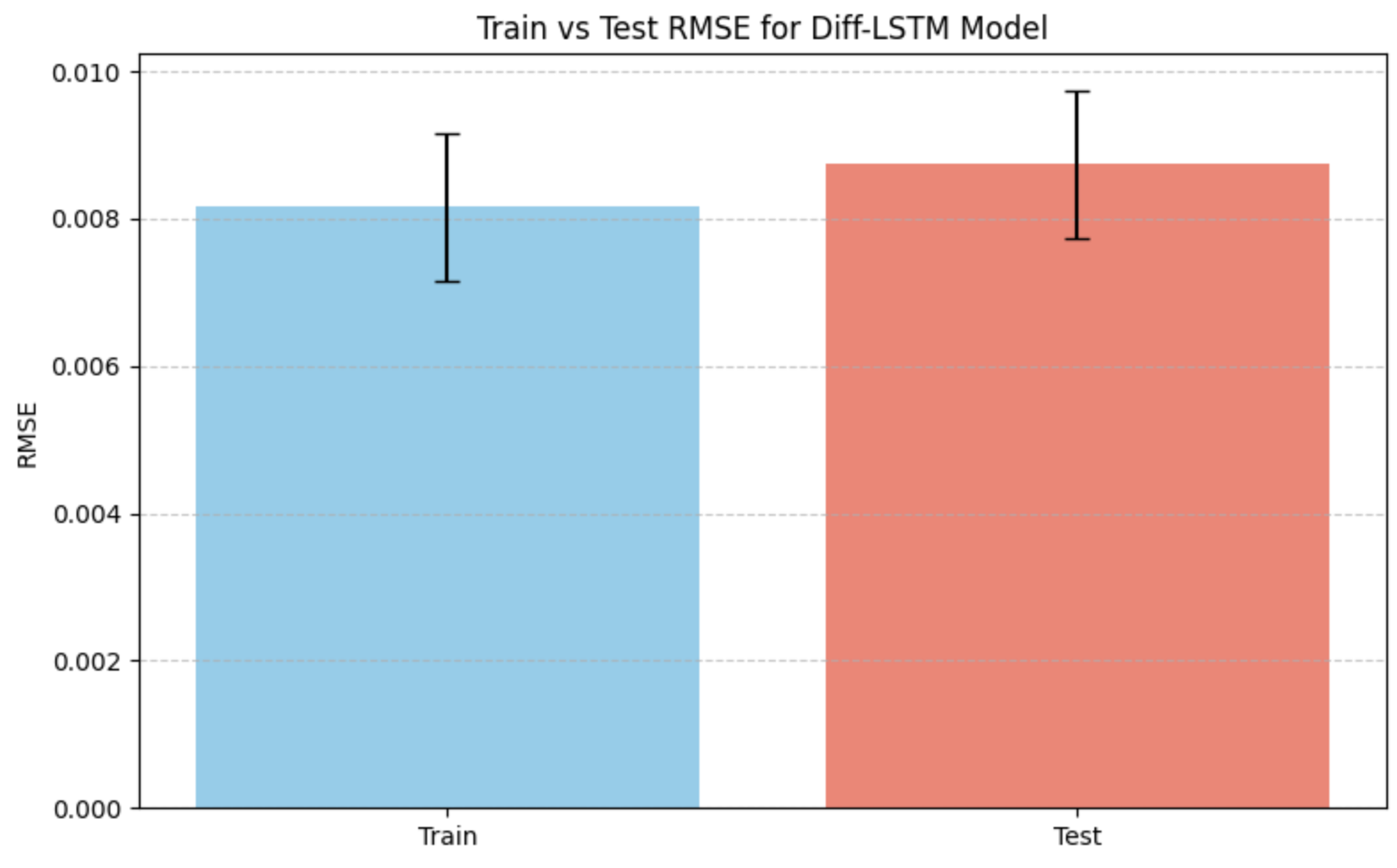}
    \includegraphics[width=0.6\textwidth]{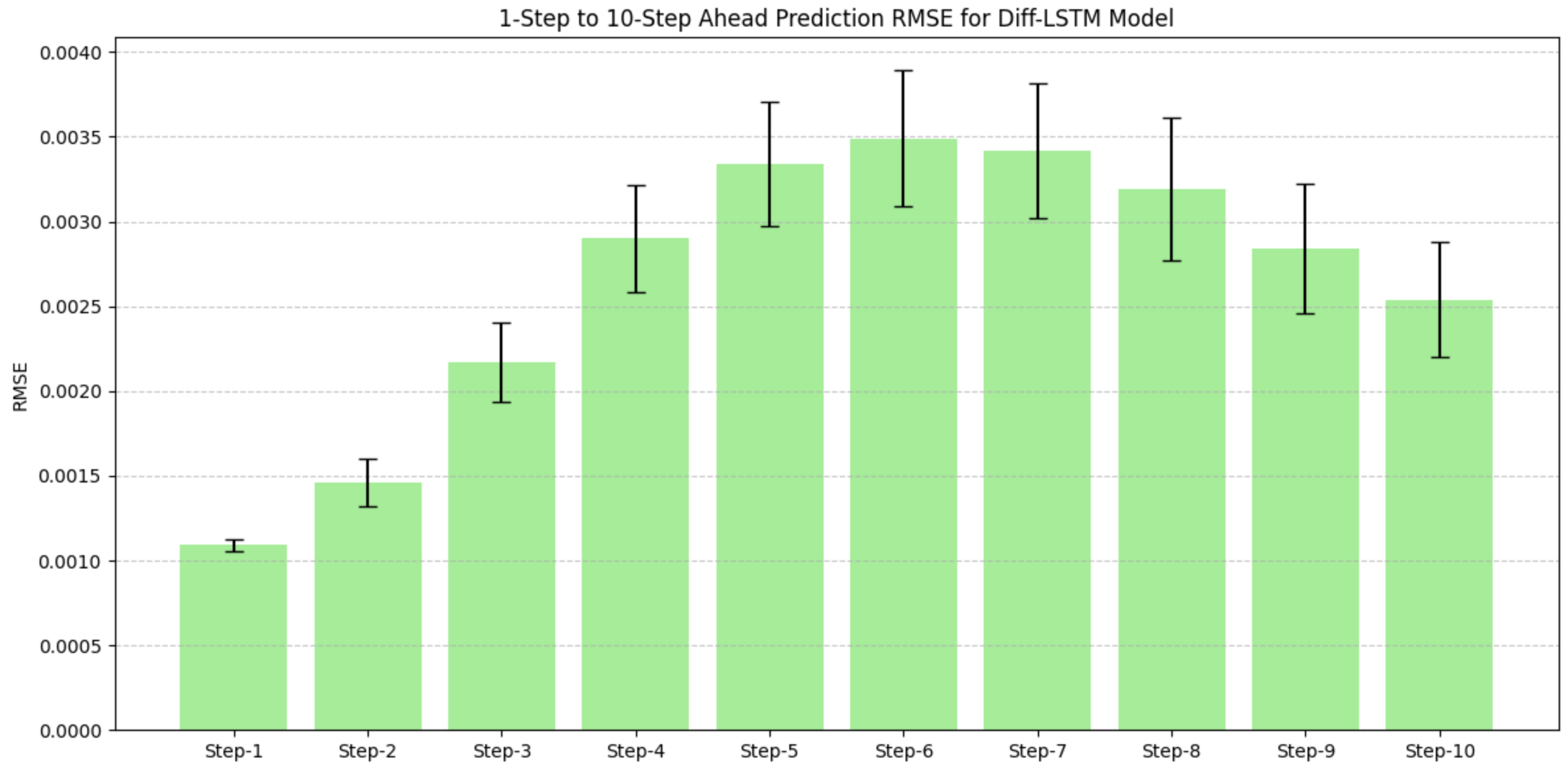}
    \caption{Mackey-Glass differential series prediction}
    \label{fig:MG_diff_RMSE}
\end{figure}

\begin{figure}
    \centering
    \includegraphics[width=0.49\textwidth]{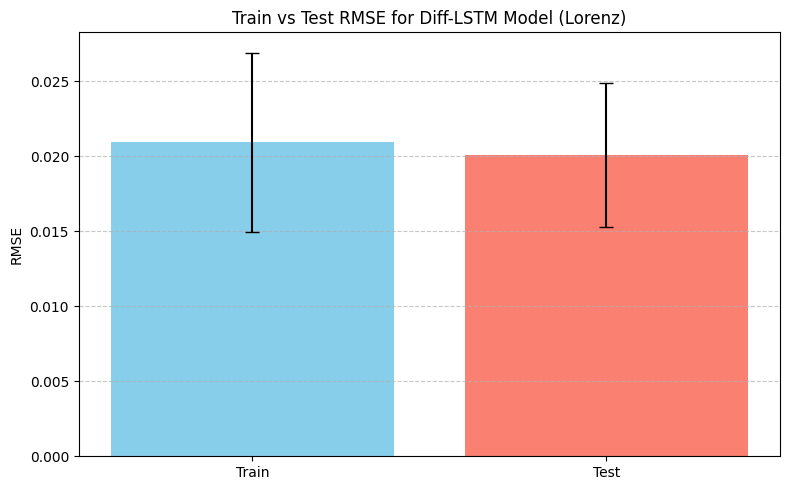}
    \includegraphics[width=0.49\textwidth]{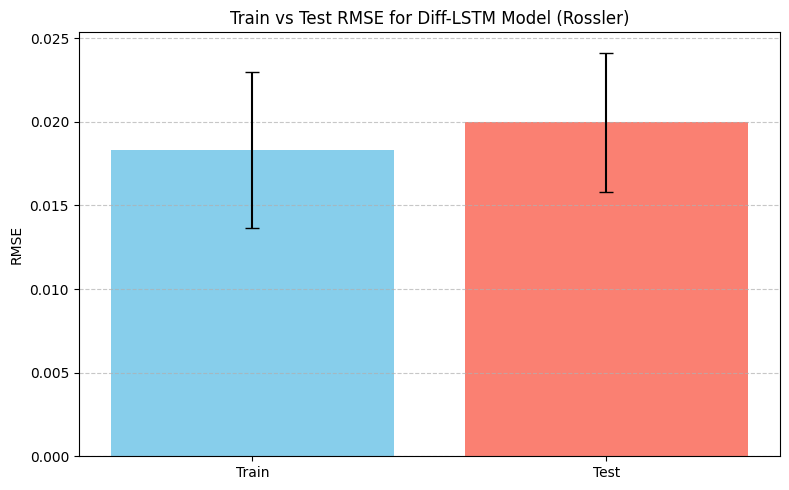}
    \includegraphics[width=0.49\textwidth]{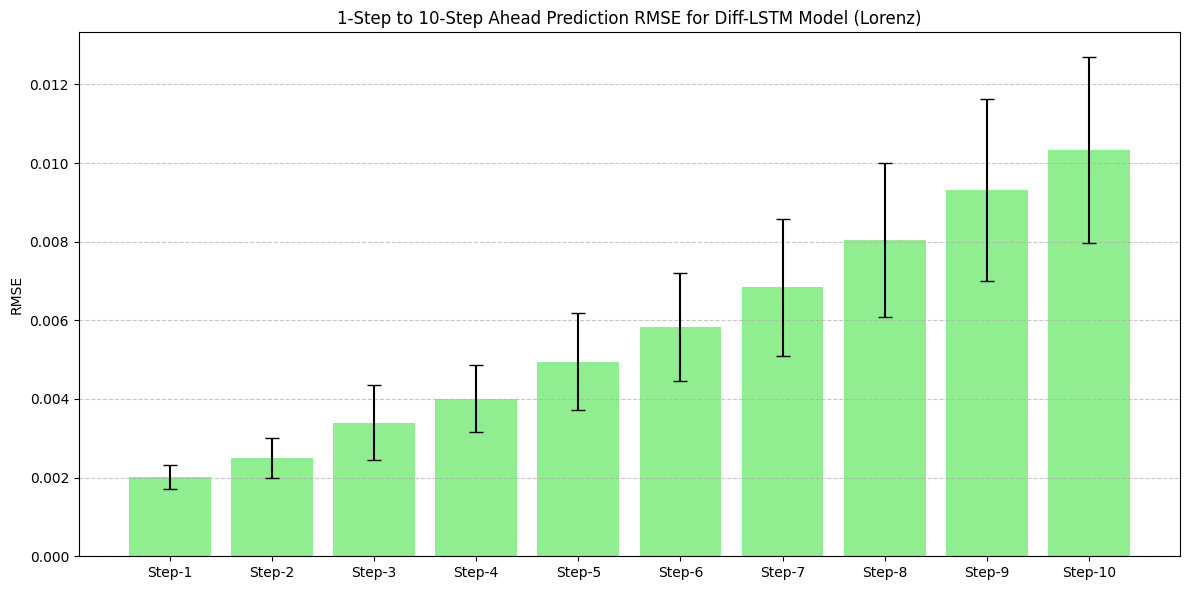}
    \includegraphics[width=0.49\textwidth]{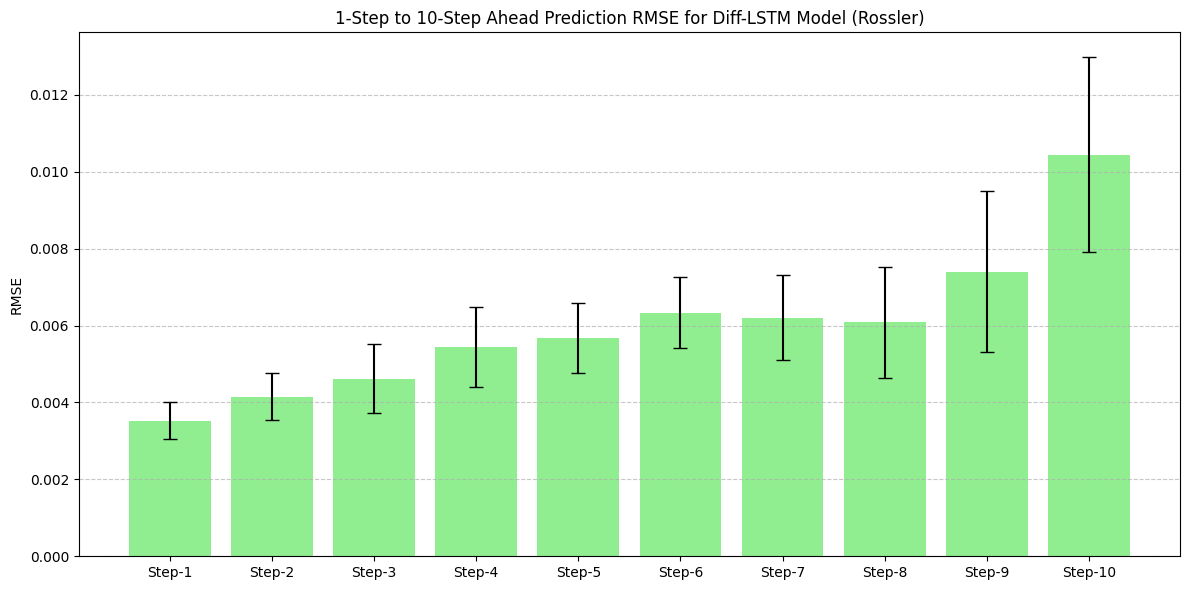}
    \caption{Lorenz and Rössler differential series prediction}
    \label{fig:Lorenz_Rossler_diff_RMSE}
\end{figure}

Furthermore, in Figures \ref{fig:MG_diff_RMSE} and \ref{fig:Lorenz_Rossler_diff_RMSE}, we present the RMSE in two different ways. First, we compare the training and testing RMSE across 10 prediction horizons. Second, we plot the RMSE for 1 to 10 step ahead predictions. It is important to note that in the results presented here, we have set \(\lambda = 1\), meaning we have assigned equal weights to the loss function for both the original and differential series.  

Interestingly, for the Mackey-Glass differential series, long-horizon predictions (8 to 10 step ahead forecasts) show a decrease in RMSE values, as illustrated in Figure \ref{fig:MG_diff_RMSE}. This can be attributed to the influence of the delay term on the differential series— as the prediction steps approach the delay term, predictive power increases. This argument is further supported by the fact that such behavior is not observed in the Lorenz and Rössler series.

\begin{figure}
    \centering
    \includegraphics[width=0.7\textwidth]{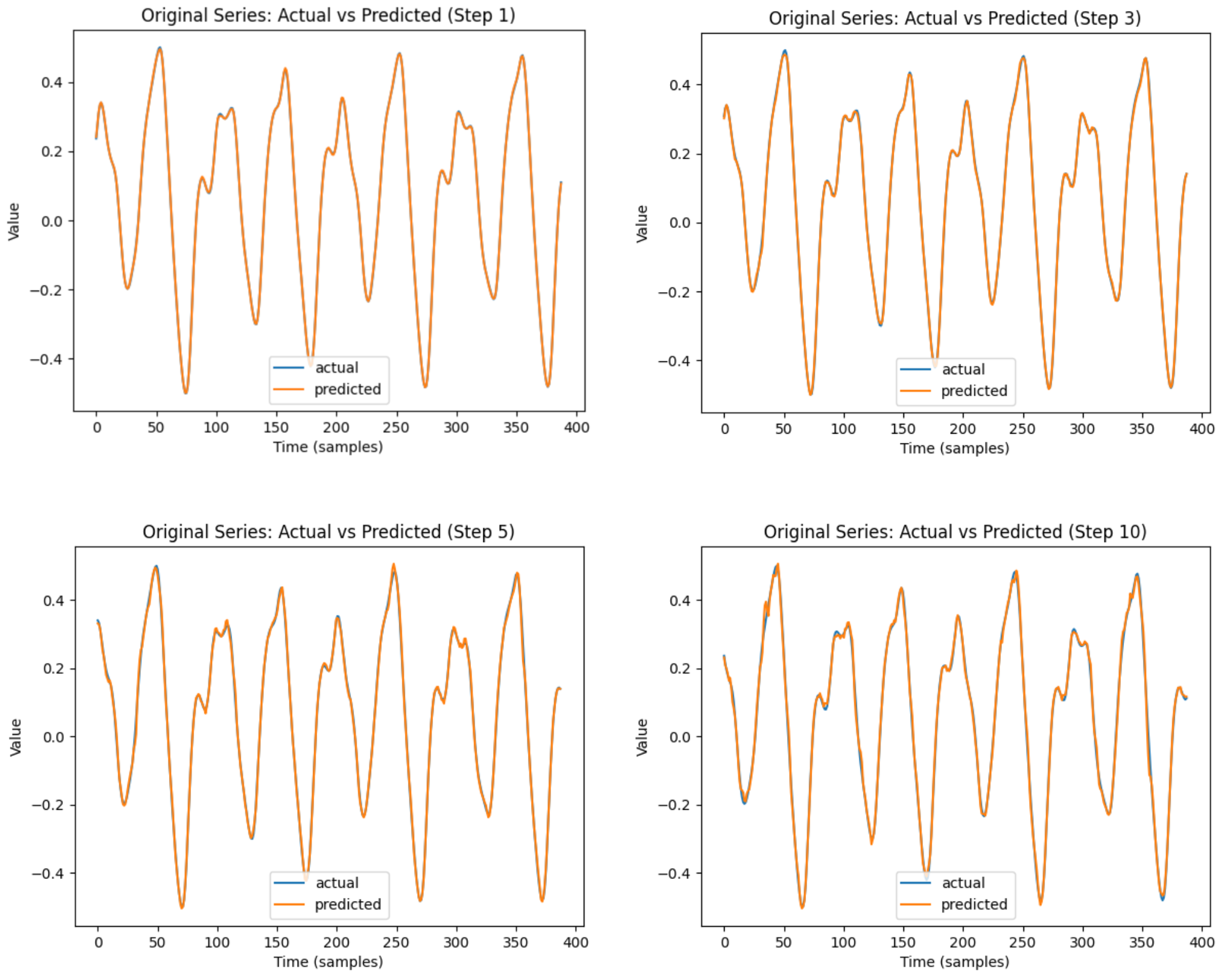}
    \includegraphics[width=0.7\textwidth]{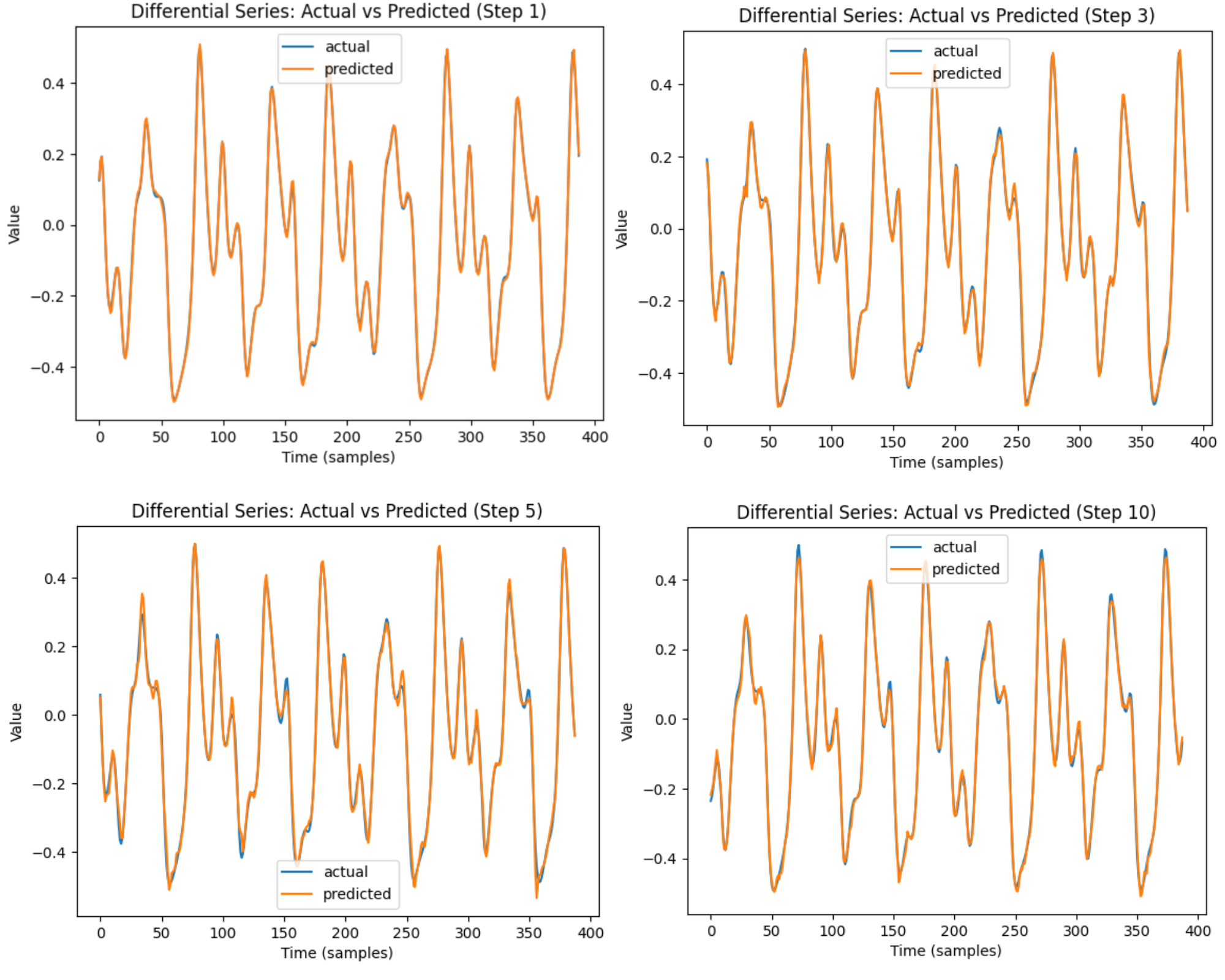}
    \caption{Actual versus predicted Mackey-Glass series for short and long horizons. }
    \label{fig:series_visualisation}
\end{figure}

\begin{figure}
    \centering
    \includegraphics[width=0.7\textwidth]{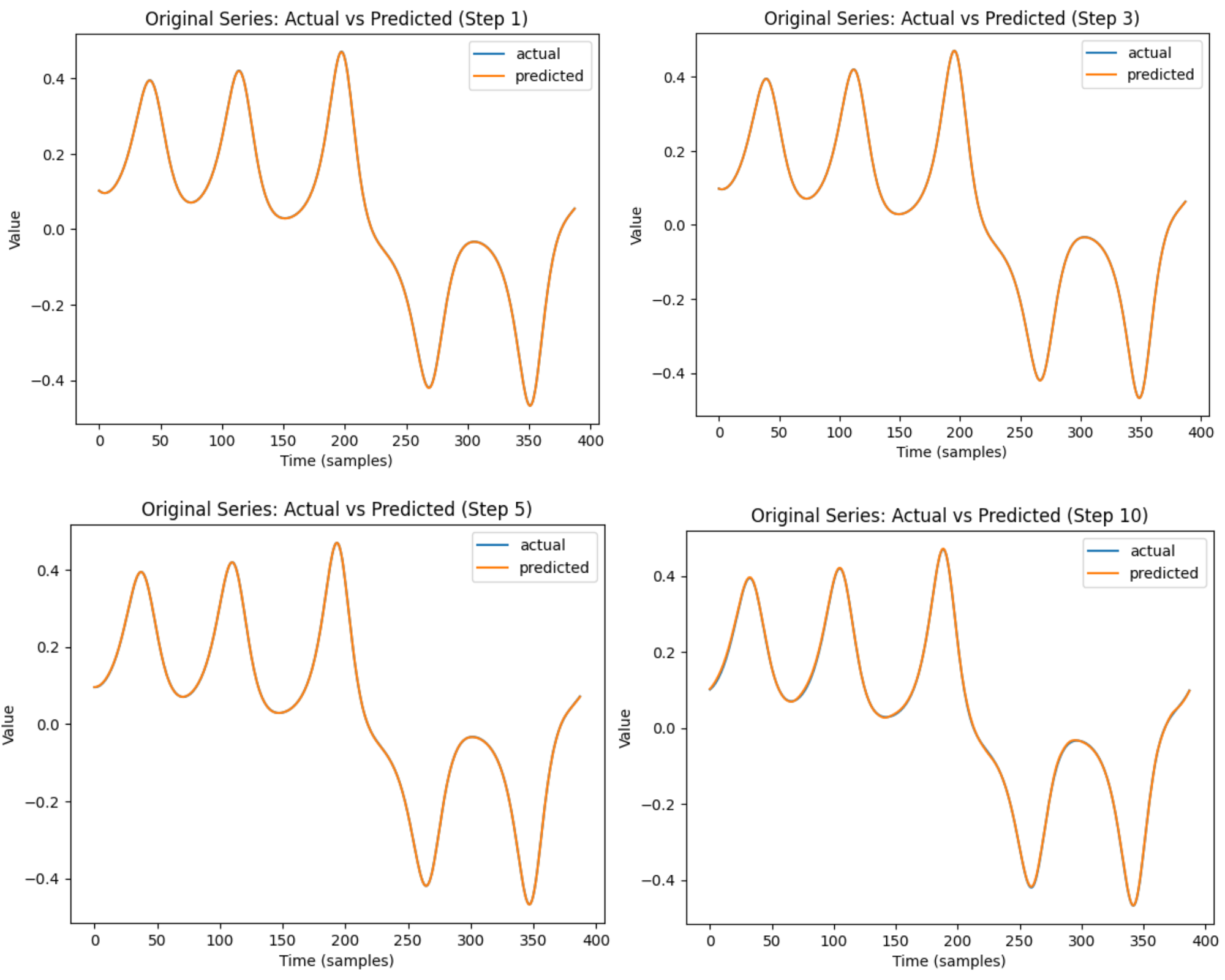}
    \includegraphics[width=0.7\textwidth]{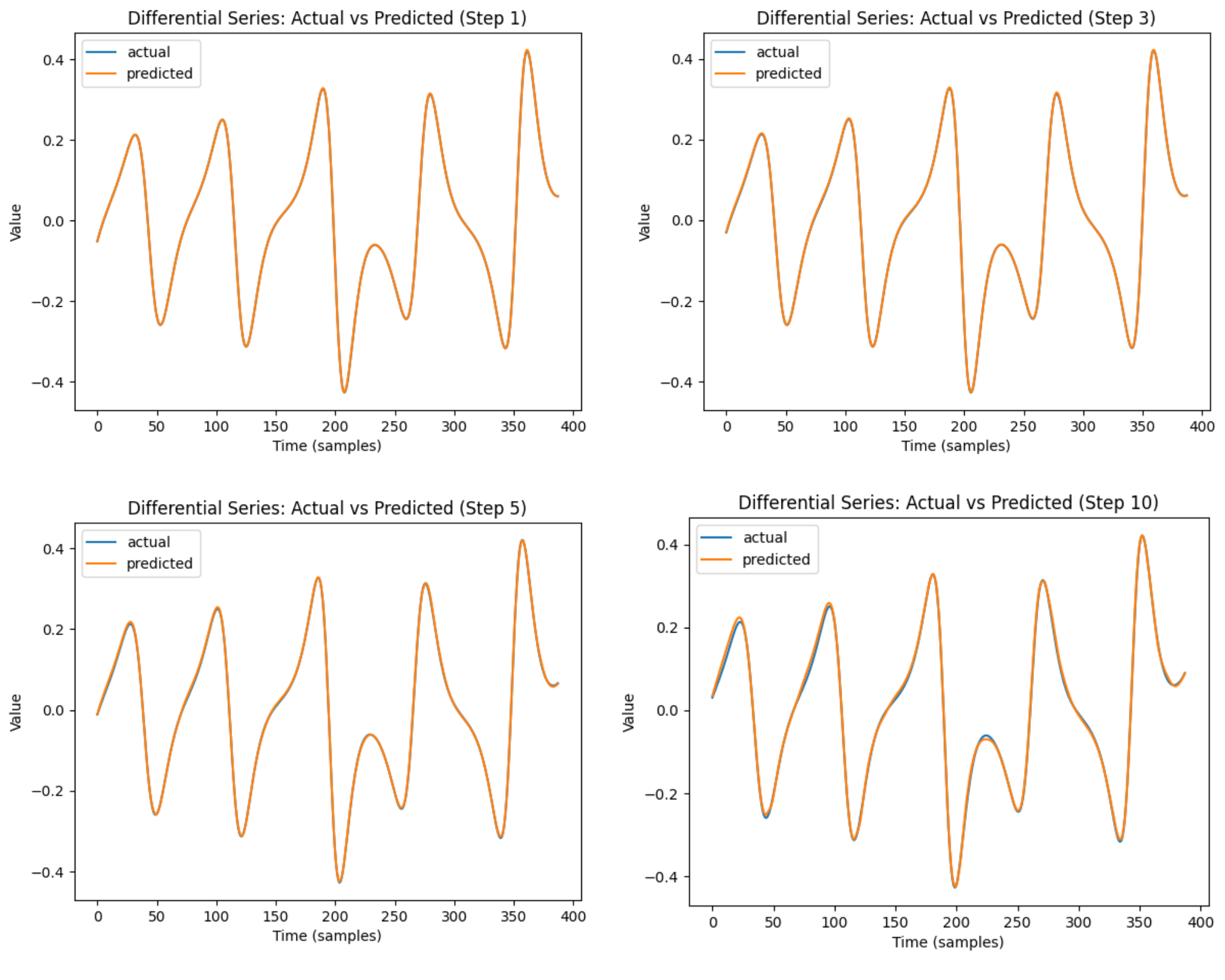}
    \caption{Actual versus predicted Lorenz series for short and long horizons}
    \label{fig:lorenz_series_visualisation}
\end{figure}

\begin{figure}
    \centering
    \includegraphics[width=0.7\textwidth]{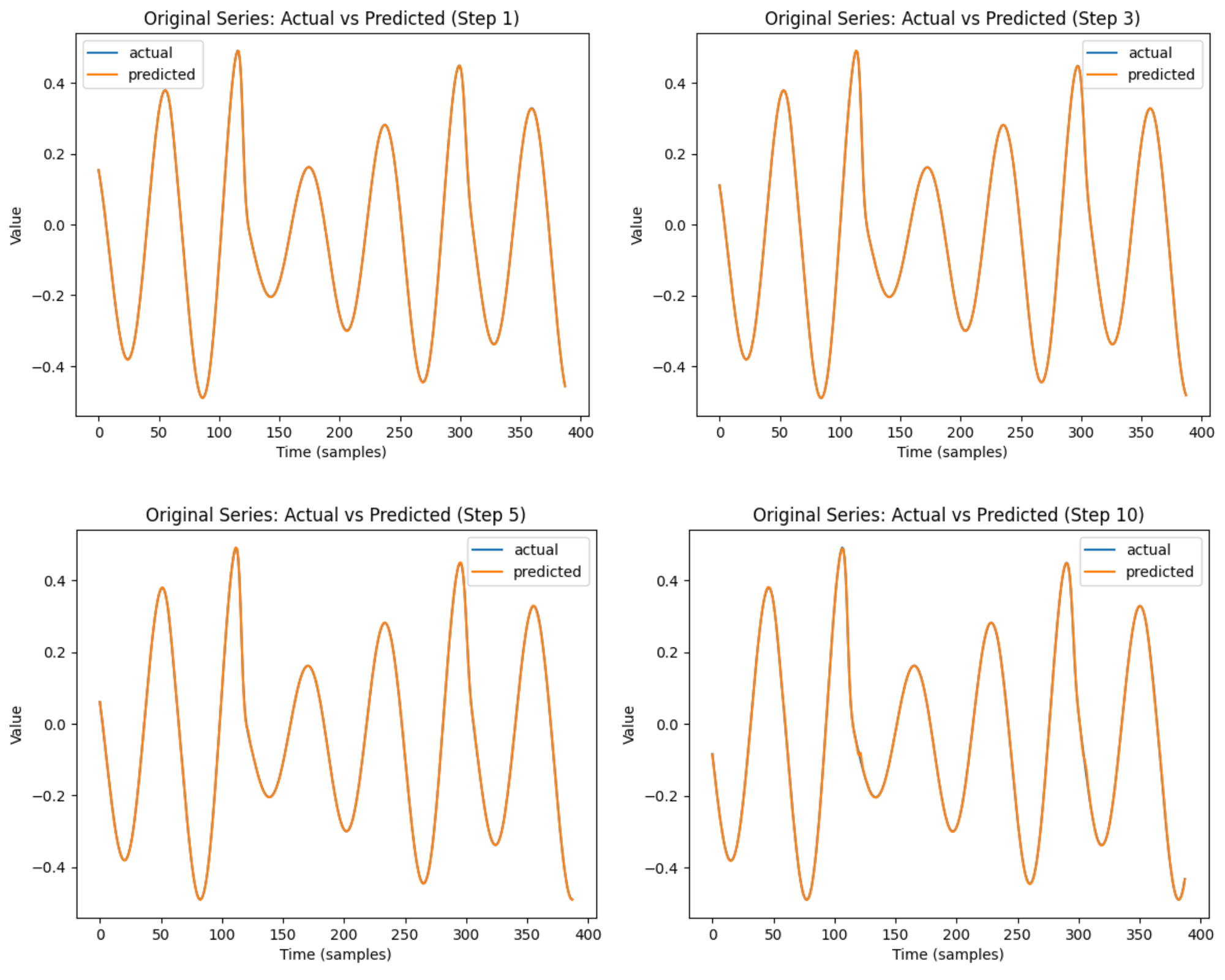}
    \includegraphics[width=0.7\textwidth]{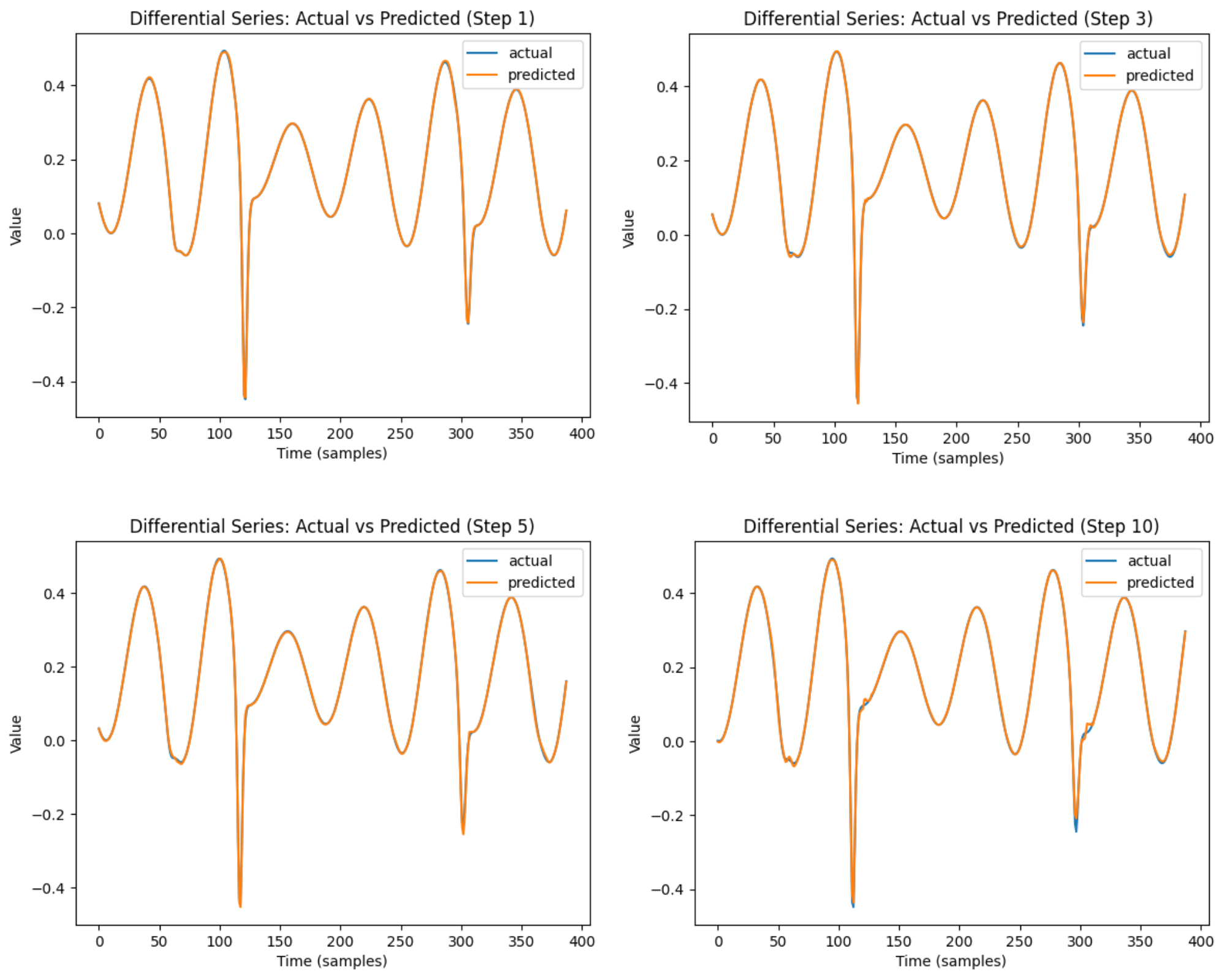}
    \caption{Actual versus predicted Rössler series for short and long horizons}
    \label{fig:Rossler_series_visualisation}
\end{figure}

For both the original and differential series in our simulated cases, a visual representation of the actual versus predicted values for specific time horizons is shown in Figures \ref{fig:series_visualisation}, \ref{fig:lorenz_series_visualisation}, and \ref{fig:Rossler_series_visualisation}. A close comparison between the results from \cite{access2021} and those obtained here using Diff-LSTM reveals an improved predictive performance of Diff-LSTM, particularly in instances of significant shape changes. This improvement highlights the impact of differential learning and its role in enhancing prediction accuracy.  

A novel feature of our architecture is its ability to simultaneously predict both the time series and its corresponding differential series, provided we have a sound theoretical basis for computing these differentials. As discussed in Section \ref{sec: ACI_results}, when these differentials are merely estimates, they still contribute value to the predictive results of the original series, given that the hyperparameter is carefully chosen.

\paragraph{Effects of Stationarity}

For certain parameter values of our benchmark time series, such as a low \(\tau\) corresponding to small delays in the case of Mackey-Glass, the system can exhibit stationarity. In such cases, we suspect that a standard feedforward network should perform equally well in predicting future values. However, this does not diminish the applicability of differential learning, where prediction performance is enhanced by incorporating the differential series as an additional input, as demonstrated in \cite{huge2020differential}.  

For parameter ranges where the time series is non-stationary, our modified framework with the Diff-LSTM network should be particularly useful, as the memory cells in LSTM help capture long-term dependencies. This unique characteristic of LSTM makes it highly effective in modeling complex temporal dynamics, such as those observed in non-stationary time series. This assertion is strongly supported by the results on feedforward networks presented in \cite{access2021}, as well as in Tables \ref{tab:MG_Diff_LSTM_OG_pred}, \ref{tab:Lorenz_Diff_LSTM_OG_pred}, and \ref{tab:Rossler_Diff_LSTM_OG_pred}.  

\subsection{Empirical application}

\subsubsection{ACI-Finance time series}
\label{sec: ACI_results}

\begin{figure}
    \centering
    \includegraphics[width=0.6\textwidth]{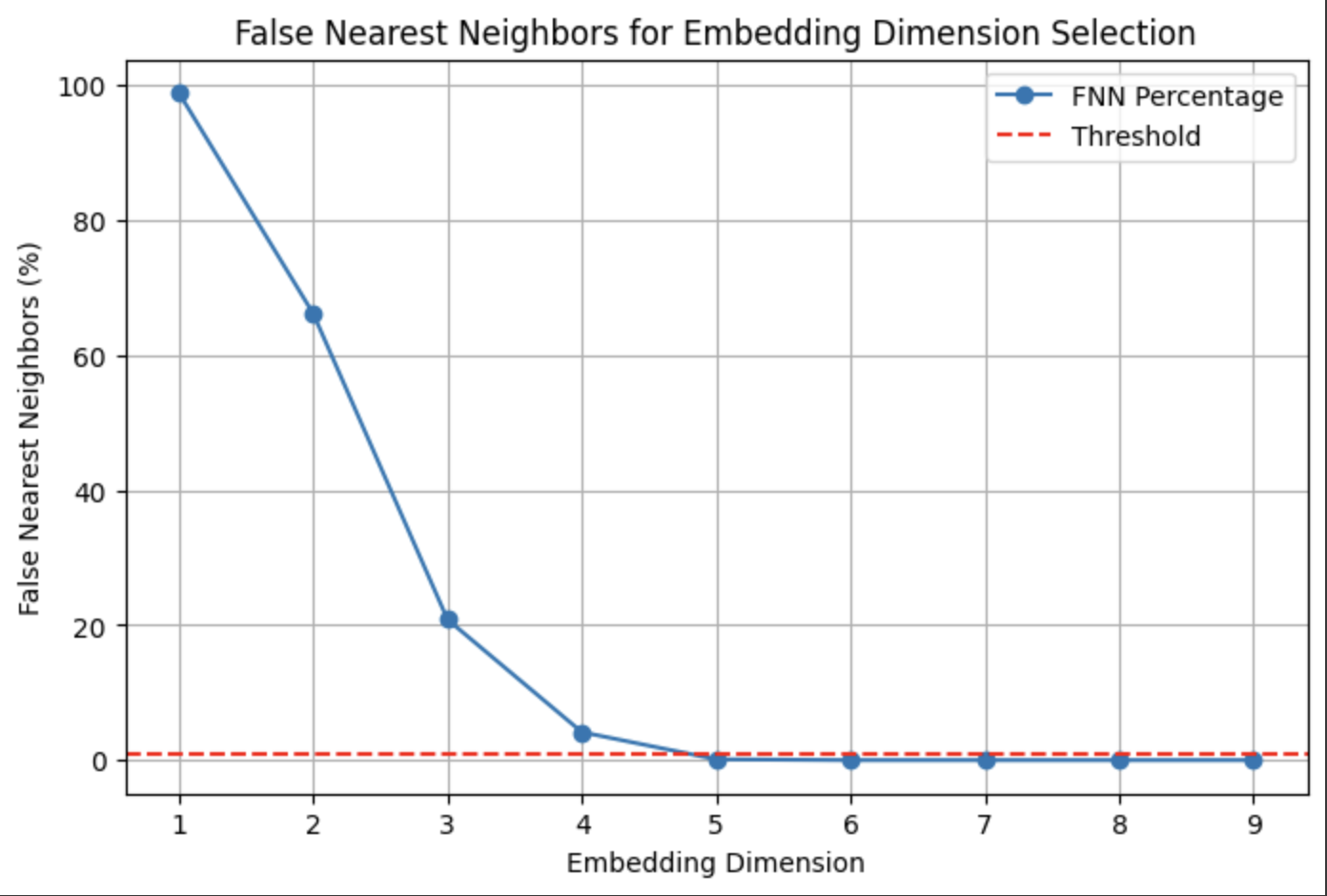}
    \caption{Optimal embedding dimensions for the ACI-Finance time series}
    \label{fig:FalseNearestNeighbor}
\end{figure}

The ACI-Finance time series consists of daily closing prices of ACI Worldwide Inc., a company listed on the NASDAQ stock exchange  
\url{http://www.nasdaq.com/symbol/aciw/stock-chart}. The dataset includes closing stock prices from December 2006 to February 2010, totaling approximately 800 data points. Our first task is to determine the embedding dimension \(D\) for this time series. As mentioned in Section \ref{sec: Taken}, this can be achieved using the false nearest neighbors algorithm \cite{RevModPhys.65.1331}, as shown in Figure \ref{fig:FalseNearestNeighbor}. Based on this analysis, we conclude that \(D = 5\). The values for the time lag and prediction horizon are kept the same as in our benchmark test cases.  

An important consideration in our approach is the estimation of differentials. This is a non-trivial problem because poor estimates introduce noise, leading to degraded model performance. Many basic methods for estimating derivatives, such as first differences, are unsuitable for real-world problems due to their sensitivity to noise. However, we can significantly improve our results by employing more sophisticated techniques to obtain differential values and by selecting optimal hyperparameters in the regularized loss function. To achieve this, we use the Savitzky-Golay filter \cite{savitzky-golay}.  

The Savitzky-Golay filter operates by fitting a low-degree polynomial to a moving window of data points using the method of least squares. For each position in the time series, the filter fits a polynomial to the data within a window centered at that point. The derivative of this polynomial at the central point is then taken as the estimated differential value for that position. The key hyperparameters of this filter include the window size (the number of data points included in each local fit) and the polynomial order (the degree of the polynomial used for fitting). A higher-order polynomial can capture more complex features but may overfit noise if not properly tuned. We find that a window size of 5 and a polynomial order of 3 work well for our case.  

Another crucial hyperparameter requiring tuning is the differential regularization parameter \(\lambda\). To determine the optimal hyperparameter values, we employ a simple grid search, and we find that \(\lambda = 1/9\) yields the best results. Other effective search techniques, such as Bayesian optimization methods, as for instance the tree-structured Parzen estimator proposed by Bergstra et al. \cite{NIPS2011_86e8f7ab}, can perform well in specific use cases within a machine learning pipeline.  

It is important to note that this choice of derivative estimation is specific to this problem. However, within our framework, prediction tasks using neural networks often reduce to finding an appropriate derivative estimate of the series. In other cases, more advanced techniques, such as smoothing splines, may be necessary. A well-chosen derivative estimation method ensures a stable training process across a wide range of \(\lambda\).

\subsubsection{Results and discussion for ACI-Finance time series}

\begin{figure}
    \centering
    \includegraphics[width=0.6\textwidth]{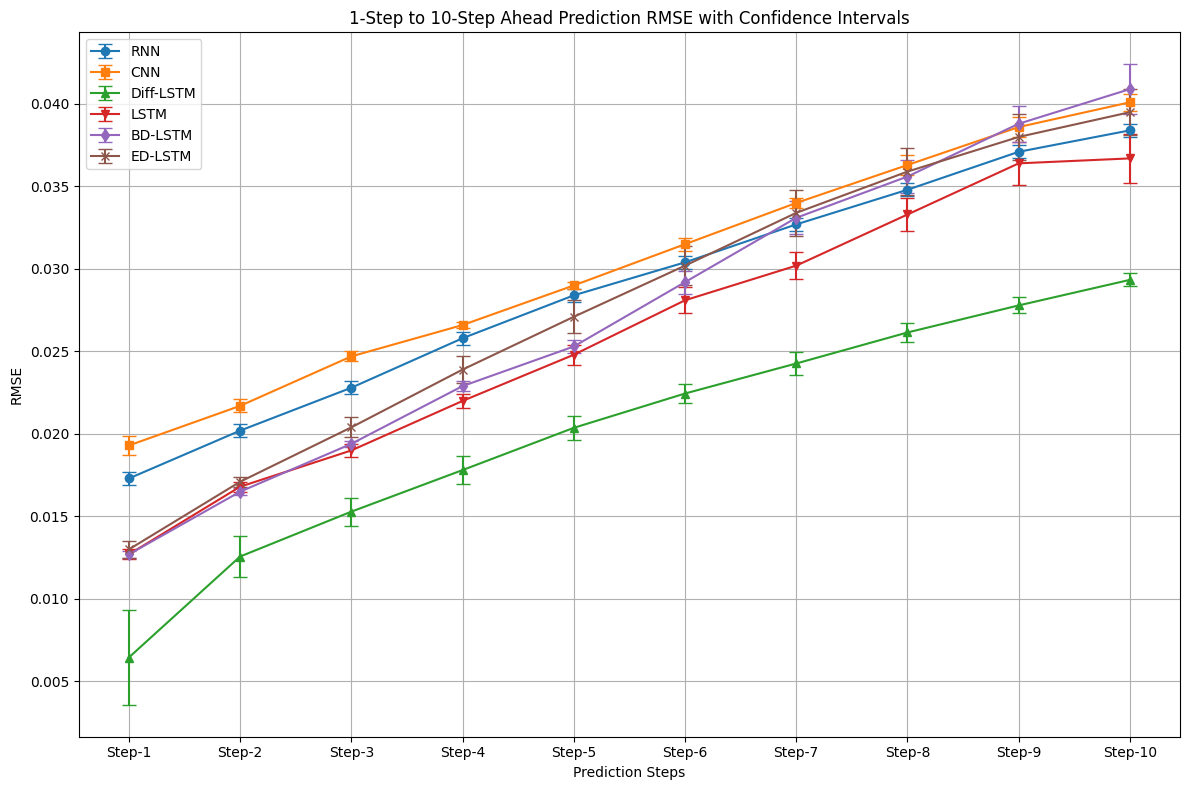}
    \caption{1 to 10 step ahead prediction for the ACI-Finance}
    \label{fig:ACI_10_step_prediction}
\end{figure}

\begin{figure}
    \centering
    \includegraphics[width=0.6\textwidth]{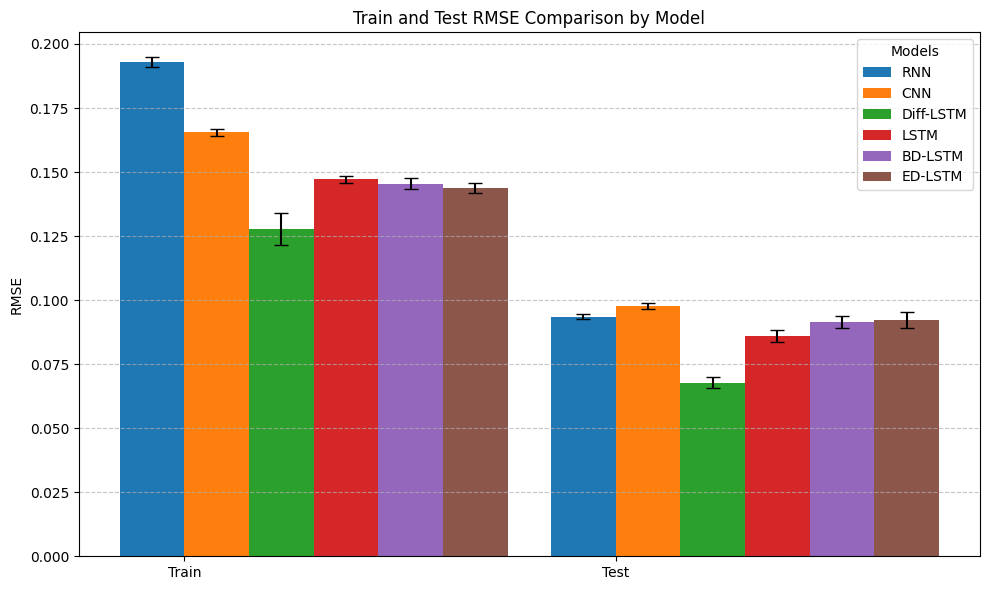}
    \caption{RMSE across 10 prediction horizons for the ACI-Finance}
    \label{fig:ACI_Train_Test_RMSE}
\end{figure}

As shown in Table \ref{tab: ACI_Finance_RMSE}, the testing performance surpasses the training performance, and the prediction error (RMSE) increases with the time horizon. Once again, Diff-LSTM outperforms other deep neural network architectures. Figure \ref{fig:ACI_10_step_prediction} compares the mean and 95\% confidence interval of the RMSE across different prediction horizons for various architectures. A similar comparison between the training and testing performance of different deep neural network architectures is presented in Figure \ref{fig:ACI_Train_Test_RMSE}.  

\begin{figure}
    \centering
    \includegraphics[width=0.7\textwidth]{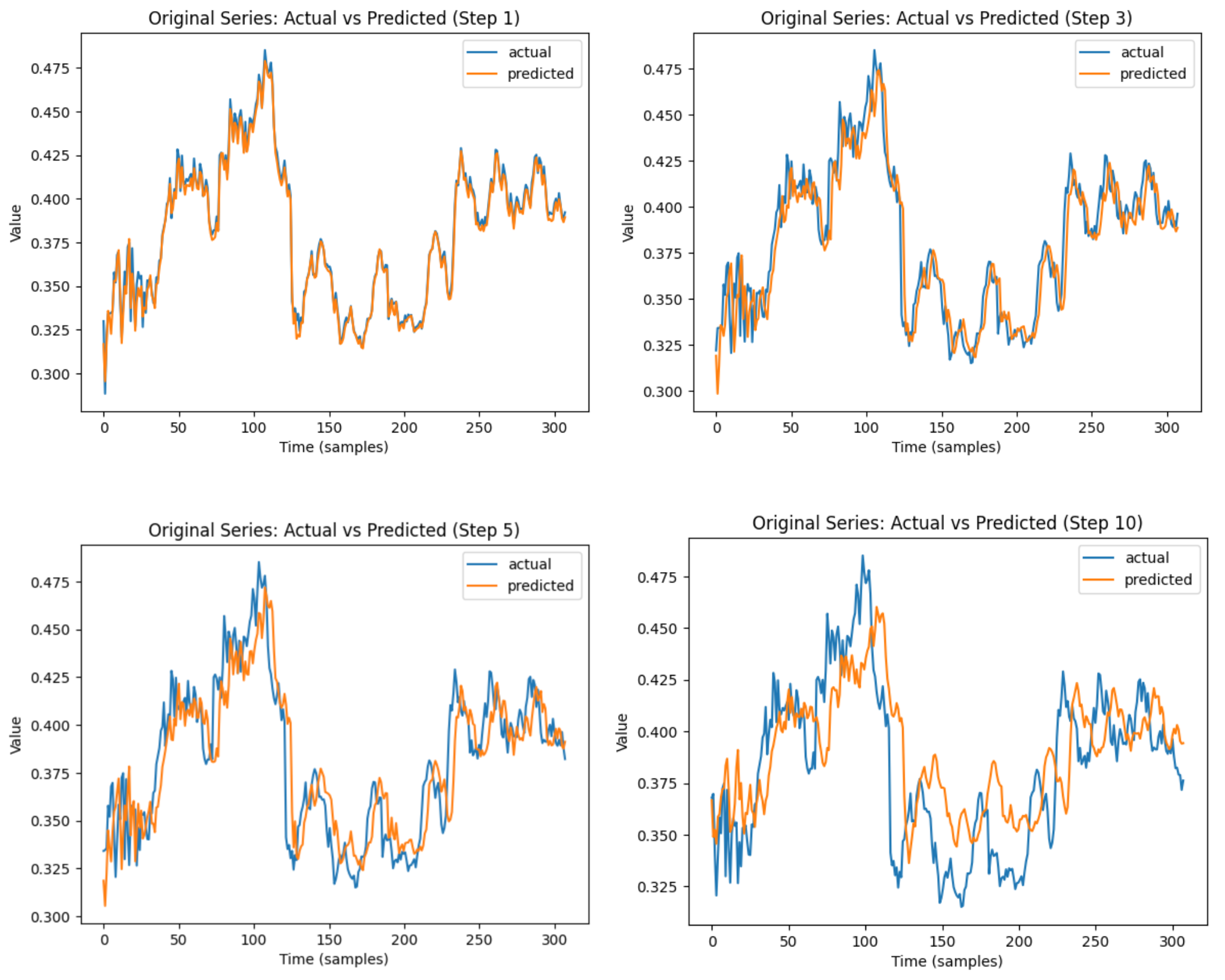}
    \caption{Actual versus predicted for ACI-Finance series for short and long horizons.}
    \label{fig:ACI_series_visualisation}
\end{figure}

Figure \ref{fig:ACI_series_visualisation} illustrates the ACI-Finance prediction performance of Diff-LSTM for selected prediction horizons, highlighting how prediction accuracy deteriorates as the time horizon increases. This trend contrasts sharply with the results for the simulated series, shown in Section \ref{sec:simulated_results}, where differences between actual and predicted values (shown in blue and orange, respectively) are almost imperceptible over longer prediction horizons, as seen in Figures \ref{fig:series_visualisation}, \ref{fig:lorenz_series_visualisation}, and \ref{fig:Rossler_series_visualisation}. The primary reason for this difference lies in the accuracy of our differential estimates. While the Savitzky-Golay filter remains a widely used method for polynomial fitting, more sophisticated differential estimation techniques exist that could potentially enhance accuracy. Exploring these advanced methods may improve prediction performance and mitigate the limitations observed at longer prediction horizons. However, it is important to recognize that no universally applicable method for differential estimation exists, as the optimal approach varies significantly across different disciplines.  

This pattern aligns with the visual comparisons of the best models for different series presented in \cite{access2021}, reinforcing the idea that although Diff-LSTM provides better predictions even when differentials are estimated, its performance remains constrained by the quality of those estimates.  

By carefully selecting the regularization hyperparameter, we can achieve optimal predictive performance for real-world datasets. Augmenting the dataset with differential estimates and fine-tuning the model accordingly allows us to focus on the primary objective: achieving the best possible predictions for the original time series.

\begin{table}
\centering
\caption{Mean and 95\% confidence interval of the RMSE for the ACI-Finance prediction.}
\label{tab: ACI_Finance_RMSE}
\resizebox{\textwidth}{!}{
\begin{tabular}{@{}lcccccc@{}}
\toprule
        & \textbf{RNN}            & \textbf{CNN}            & \textbf{Diff-LSTM}      & \textbf{LSTM}           & \textbf{BD-LSTM}        & \textbf{ED-LSTM}        \\ \midrule
\textbf{Train} & 0.1930 ± 0.0018 & 0.1655 ± 0.0013 & \textbf{0.1278 ± 0.0064}          & 0.1471 ± 0.0014 & 0.1454 ± 0.0021 & 0.1437 ± 0.0019 \\
\textbf{Test}  & 0.0936 ± 0.0009 & 0.0978 ± 0.0011 & \textbf{0.0677 ± 0.0021}          & 0.0860 ± 0.0025 & 0.0915 ± 0.0023 & 0.0923 ± 0.0032 \\
\textbf{Step-1} & 0.0173 ± 0.0004 & 0.0193 ± 0.0006 & \textbf{0.0064 ± 0.0029}          & 0.0127 ± 0.0003 & 0.0127 ± 0.0002 & 0.0130 ± 0.0005 \\
\textbf{Step-2} & 0.0202 ± 0.0004 & 0.0217 ± 0.0004 & \textbf{0.0126 ± 0.0013}          & 0.0168 ± 0.0003 & 0.0165 ± 0.0002 & 0.0171 ± 0.0003 \\
\textbf{Step-3} & 0.0228 ± 0.0003 & 0.0247 ± 0.0003 & \textbf{0.0153 ± 0.0008}          & 0.0190 ± 0.0004 & 0.0194 ± 0.0002 & 0.0204 ± 0.0006 \\
\textbf{Step-4} & 0.0258 ± 0.0004 & 0.0266 ± 0.0002 & \textbf{0.0178 ± 0.0009}          & 0.0220 ± 0.0004 & 0.0229 ± 0.0003 & 0.0239 ± 0.0008 \\
\textbf{Step-5} & 0.0284 ± 0.0004 & 0.0290 ± 0.0002 & \textbf{0.0204 ± 0.0007}          & 0.0248 ± 0.0006 & 0.0253 ± 0.0004 & 0.0271 ± 0.0010 \\
\textbf{Step-6} & 0.0304 ± 0.0004 & 0.0315 ± 0.0004 & \textbf{0.0224 ± 0.0006}          & 0.0281 ± 0.0008 & 0.0292 ± 0.0007 & 0.0302 ± 0.0012 \\
\textbf{Step-7} & 0.0327 ± 0.0004 & 0.0340 ± 0.0003 & \textbf{0.0243 ± 0.0007}          & 0.0302 ± 0.0008 & 0.0331 ± 0.0010 & 0.0334 ± 0.0014 \\
\textbf{Step-8} & 0.0348 ± 0.0004 & 0.0363 ± 0.0006 & \textbf{0.0262 ± 0.0005}          & 0.0333 ± 0.0010 & 0.0356 ± 0.0010 & 0.0359 ± 0.0014 \\
\textbf{Step-9} & 0.0371 ± 0.0003 & 0.0386 ± 0.0006 & \textbf{0.0278 ± 0.0004}          & 0.0364 ± 0.0013 & 0.0388 ± 0.0011 & 0.0380 ± 0.0014 \\
\textbf{Step-10} & 0.0384 ± 0.0003 & 0.0401 ± 0.0005 & \textbf{0.0293 ± 0.0004}          & 0.0367 ± 0.0015 & 0.0409 ± 0.0015 & 0.0395 ± 0.0014 \\ \bottomrule
\end{tabular}
}
\end{table}

\paragraph{NB: Deep neural network architecture comparison}
\label{sec: comparison}

Comparing different neural network architectures is not a straightforward task. Prediction results vary based on factors such as the number of hidden layers, the choice of optimizer, and the duration of training. To ensure a fair comparison, we have used the model parameters and datasets from Chandra et al. \cite{access2021} as our baseline. By maintaining the same configurations as \cite{access2021}, we isolate the impact of architectural differences while minimizing the influence of extraneous variables. This approach enhances the validity of our comparisons, leading to more reliable and meaningful insights into the performance and capabilities of different neural network architectures.

\paragraph{Software and Data}
\label{sec:sotware_data}

An open-source Python implementation, accompanied by the corresponding data, is made available for further research. \footnote{\url{https://github.com/akashyadav08/Diff_LSTM}}

\section{Conclusion and Future work}
\label{sec:conclusion}

Although differential machine learning has recently been applied in areas like trust prediction in signed social networks \cite{Abadeh2023}, the Delta LSM (Least Squares Monte Carlo) method \cite{lind2024gas} and model calibration, its use in time series prediction tasks remains largely unexplored. In this work, we introduced a novel differential LSTM (Diff-LSTM) network that enhances time series prediction by incorporating differential learning. Our approach enables the model to process both the original time series and its differentials simultaneously, resulting in superior performance in both short-term and long-horizon forecasting. Through extensive experimentation on benchmark chaotic systems and real-world financial data, we demonstrate that Diff-LSTM outperforms conventional deep learning architectures such as CNNs, RNNs, and Encoder-Decoder LSTMs. These findings confirm that differential learning can serve as a foundational tool for improving prediction tasks across various domains.  

A key contribution of our work is the comparative analysis of model architectures, emphasizing structural differences rather than focusing solely on predictive performance. While LSTM-based architectures, as supported by previous studies \cite{pmlr-v37-jozefowicz15}, are particularly well-suited for time series forecasting, other architectures can also be adapted to integrate differential learning. This flexibility opens new possibilities for enhancing time series modeling across diverse applications. Furthermore, expanding our model’s capacity could yield even better predictive results in cases where achieving optimal forecasting accuracy is the primary objective.  

For future work, incorporating differential PCA, as introduced by Huge and Savine \cite{Huge2021Axes}, as an encoding mechanism within an Encoder-Decoder LSTM framework could further improve feature extraction and representation of complex time series data.

\section*{Acknowledgments}

This work was supported by Ayudas Fundacion BBVA a Proyectos de Investigaci\'on Cient\'ifica 2021 (Ref. Fundación BBVA-EI-2022-G. Lugosi). The authors would like to thank Dr. Antoine Savine (Hudson River Trading), Dr. Andre Souza (ESADE), Nina McClure (Reserve Bank of Australia, Bank for International Settlements) and Elain B. Balderas (Novartis) for their valuable insights and discussion.

\bibliographystyle{unsrt}  
\bibliography{references}

\begin{thebibliography}{10}

\bibitem{huge2020differential}
B.~Huge and A.~Savine.
\newblock Differential machine learning.
\newblock SSRN: https://ssrn.com/abstract=3591734, 2020.

\bibitem{HochSchm97}
S.~Hochreiter and J.~Schmidhuber.
\newblock Long short-term memory.
\newblock {\em Neural Computation}, 9(8):1735--1780, 1997.

\bibitem{MG}
M.C. Mackey and L.~Glass.
\newblock Oscillation and chaos in physiological control systems.
\newblock {\em Science}, 197(4300):287--289, 1977.

\bibitem{LorenzDeterministicNonperiodicFlow}
E.N. Lorenz.
\newblock Deterministic nonperiodic flow.
\newblock {\em Journal of Atmospheric Sciences}, 20(2):130--141, 1963.

\bibitem{Rossler1976397}
O.E. Rössler.
\newblock An equation for continuous chaos.
\newblock {\em Physics Letters A}, 57(5):397--398, 1976.

\bibitem{access2021}
R.~Chandra, R.~Gupta, and S.~Goyal.
\newblock Evaluation of deep learning models for multi-step ahead time series prediction.
\newblock {\em IEEE Access}, 9:83105--83123, 2021.

\bibitem{SALGADO2022498}
P.A. Salgado and T.P.A Perdicoúlis.
\newblock A fuzzy derivative model approach to time-series prediction⁎⁎.
\newblock {\em IFAC-PapersOnLine}, 55(30):498--503, 2022.

\bibitem{RAISSI2019686}
M.~Raissi, P.~Perdikaris, and G.E. Karniadakis.
\newblock Physics-informed neural networks: A deep learning framework for solving forward and inverse problems involving nonlinear partial differential equations.
\newblock {\em Journal of Computational Physics}, 378:686--707, 2019.

\bibitem{LIU2023116500}
F.~Liu, J.~Li, and L.~Wang.
\newblock Pi-lstm: Physics-informed long short-term memory network for structural response modeling.
\newblock {\em Engineering Structures}, 292:116500, 2023.

\bibitem{savitzky-golay}
A.~Savitzky and M.~J.~E. Golay.
\newblock Smoothing and differentiation of data by simplified least squares procedures.
\newblock {\em Analytical Chemistry}, 36(8):1627--1639, 1964.

\bibitem{FinancialAN}
Preeti, R.~Bala, and R.P. Singh.
\newblock Financial and non-stationary time series forecasting using lstm recurrent neural network for short and long horizon.
\newblock {\em 2019 10th International Conference on Computing, Communication and Networking Technologies (ICCCNT)}, pages 1--7, 2019.

\bibitem{RePEc:arx:papers:2405.01233}
Pedro~Duarte Gomes.
\newblock {Mathematics of Differential Machine Learning in Derivative Pricing and Hedging}.
\newblock Papers 2405.01233, arXiv.org, May 2024.

\bibitem{multitask_learning}
R.~Caruana.
\newblock Multitask learning.
\newblock {\em Machine Learning}, 28(1):41--75, 1997.

\bibitem{Gers2000LearningTF}
F.A. Gers, J.~Schmidhuber, and F.~Cummins.
\newblock Learning to forget: Continual prediction with lstm.
\newblock {\em Neural Computation}, 12:2451--2471, 2000.

\bibitem{Kingma:2014vow}
D.~P. Kingma and J.~Ba.
\newblock Adam: A method for stochastic optimization.
\newblock arxiv:1412.6980, 2014.

\bibitem{takens1981detecting}
F.~Takens.
\newblock Detecting strange attractors in turbulence.
\newblock In D.~Rand and L.-S. Young, editors, {\em Dynamical Systems and Turbulence, Warwick 1980}, Lecture Notes in Mathematics. Springer Berlin Heidelberg, 1981.

\bibitem{frazier2004chaos}
C.~Frazier and K.~Kockelman.
\newblock Chaos theory and transportation systems: Instructive example.
\newblock {\em Transportation Research Record: Journal of the Transportation Research Board}, 20:9--17, 2004.

\bibitem{PhysRevLett.50.346}
P.~Grassberger and I.~Procaccia.
\newblock Characterization of strange attractors.
\newblock {\em Phys. Rev. Lett.}, 50:346--349, 1983.

\bibitem{sauer1991embedology}
T.~Sauer, J.~A. Yorke, and M.~Casdagli.
\newblock Embedology.
\newblock {\em Journal of statistical Physics}, 65:579--616, 1991.

\bibitem{RevModPhys.65.1331}
H.~D.~I. Abarbanel, R.~Brown, J.~J. Sidorowich, and L.~Sh. Tsimring.
\newblock The analysis of observed chaotic data in physical systems.
\newblock {\em Rev. Mod. Phys.}, 65:1331--1392, 1993.

\bibitem{Grebogifractaldimension}
C.~Grebogi, E.~Ott, and J.A. Yorke.
\newblock Chaos, strange attractors, and fractal basin boundaries in nonlinear dynamics.
\newblock {\em Science}, 238(4827):632--638, 1987.

\bibitem{Kaplan1979ChaoticBO}
J.L. Kaplan and J.A. Yorke.
\newblock Chaotic behavior of multidimensional difference equations.
\newblock In Peitgen and Walther, editors, {\em Functional Differential Equations and Approximation of Fixed Points}, volume 730 of {\em Lecture Notes in Mathematics}. Springer Berlin Heidelberg, 1979.

\bibitem{NIPS2011_86e8f7ab}
J.~Bergstra, R.~Bardenet, Y.~Bengio, and B.~K\'{e}gl.
\newblock Algorithms for hyper-parameter optimization.
\newblock In J.~Shawe-Taylor, R.~Zemel, P.~Bartlett, F.~Pereira, and K.Q. Weinberger, editors, {\em Advances in Neural Information Processing Systems}, volume~24. Curran Associates, Inc., 2011.

\bibitem{Abadeh2023}
Maryam~Nooraei Abadeh and Mansooreh Mirzaie.
\newblock A differential machine learning approach for trust prediction in signed social networks.
\newblock {\em The Journal of Supercomputing}, 79(9):9443--9466, 2023.

\bibitem{lind2024gas}
Peter~Pommerg{\aa}rd Lind.
\newblock Gas storage valuation using delta least squares monte carlo method.
\newblock {\em Available at SSRN 4831976}, 2024.

\bibitem{pmlr-v37-jozefowicz15}
R.~Jozefowicz, W.~Zaremba, and I.~Sutskever.
\newblock An empirical exploration of recurrent network architectures.
\newblock In Francis Bach and David Blei, editors, {\em Proceedings of the 32nd International Conference on Machine Learning}, volume~37 of {\em Proceedings of Machine Learning Research}, Lille, France, 2015. PMLR.

\bibitem{Huge2021Axes}
B.~Huge and A.~Savine.
\newblock Axes that matter: P{C}{A} with a difference.
\newblock {\em Risk.net}, 2021.

\end{thebibliography}

\end{document}